\documentclass[journal]{IEEEtran}
\usepackage{xurl} 
\usepackage{hyperref}
\usepackage{fix-cm}
\usepackage{amsmath,amsfonts}
\usepackage{array}
\usepackage[noadjust]{cite}
\usepackage{textcomp}
\usepackage{stfloats}
\usepackage{url}
\usepackage{verbatim}
\usepackage{graphicx}
\def\BibTeX{{\rm B\kern-.05em{\sc i\kern-.025em b}\kern-.08em
    T\kern-.1667em\lower.7ex\hbox{E}\kern-.125emX}}
\usepackage{balance}
\usepackage{tikz}
\usetikzlibrary{fadings}
\usetikzlibrary{patterns}
\usetikzlibrary{shadows.blur}
\usetikzlibrary{shapes}   
\usepackage{siunitx}
\DeclareSIUnit\mt{\milli\tesla} 
\usepackage[utf8]{inputenc}
\usepackage[english]{babel}
\usepackage{xurl} 
\usepackage[linesnumbered,ruled,lined]{algorithm2e}
\usepackage{array}
\usepackage[noadjust]{cite}
\usepackage{textcomp}
\usepackage{url}
\usepackage{verbatim}
\usepackage{graphicx}
\def\BibTeX{{\rm B\kern-.05em{\sc i\kern-.025em b}\kern-.08em
    T\kern-.1667em\lower.7ex\hbox{E}\kern-.125emX}}
\usepackage{balance}
\usepackage{multirow} 
\usepackage{setspace}
\usepackage{graphicx}
\usepackage{url}
\usepackage{subcaption}
\usepackage{amsmath}
\usepackage{algcompatible}
\usepackage{array}
\usepackage{amsthm}
\usepackage{adjustbox}
\usepackage{color}

\usepackage{pgfplots}
\usepackage{tikz}
\usetikzlibrary{fadings}
\usetikzlibrary{patterns}
\usetikzlibrary{shadows.blur}
\usetikzlibrary{shapes} 

\usepackage{bbm}
\let\oldnl\nl
\newcommand{\nonl}{\renewcommand{\nl}{\let\nl\oldnl}}
\newcolumntype{L}[1]{>{\raggedright\let\newline\\\arraybackslash\hspace{0pt}}m{#1}}
\newcolumntype{C}[1]{>{\centering\let\newline\\\arraybackslash\hspace{0pt}}m{#1}}
\newcolumntype{R}[1]{>{\raggedleft\let\newline\\\arraybackslash\hspace{0pt}}m{#1}}

\algnewcommand\KwEvl{\textbf{Evaluation:}}
\usepackage{booktabs}

\usepackage{makecell}
\usepackage{amsmath}
\usepackage{siunitx}
\usepackage{caption} 
\usepackage{etoolbox}
\usepackage{tikz}

\newrobustcmd*{\mysquare}[1]{\tikz{\filldraw[draw=#1,fill=#1] (0,0)
rectangle (0.2cm,0.2cm);}}

\newrobustcmd*{\mycircle}[1]{\tikz{\filldraw[draw=#1,fill=#1] (0,0) circle [radius=0.1cm];}}

\newrobustcmd*{\mytriangle}[1]{\tikz{\filldraw[draw=#1,fill=#1] (0,0) --
(0.2cm,0) -- (0.1cm,0.2cm);}}

\usepackage{caption}

\begin{document}


\title{Digital Twin-Assisted Explainable AI for Robust Beam Prediction in mmWave MIMO Systems}
\author{Nasir~Khan,~\IEEEmembership{Graduate Student Member,~IEEE}, Asmaa~Abdallah,~\IEEEmembership{Member,~IEEE},\\Abdulkadir~Celik,~\IEEEmembership{Senior~Member,~IEEE},~Ahmed~M.~Eltawil,~\IEEEmembership{Senior Member,~IEEE},\\ and~Sinem Coleri,~\IEEEmembership{Fellow,~IEEE}

\thanks{Nasir~Khan and Sinem Coleri are with the department of Electrical and Electronics Engineering, Koc University, Istanbul, Turkey, email: $\lbrace$nkhan20, scoleri$\rbrace$@ku.edu.tr. This work is supported by Scientific and Technological Research Council of Turkey Grant $\#$119C058 and Ford Otosan.}
 \thanks{Asmaa Abdallah, Abdulkadir Celik, and Ahmed M. Eltawil are with the Computer, Electrical, and Mathematical Sciences and Engineering Division, King Abdullah University of Science and Technology, Thuwal 23955, Saudi Arabia (e-mail:  asmaa.abdallah@kaust.edu.sa; abdulkadir.celik@kaust.edu.sa;  ahmed.eltawil@kaust.edu.sa)}}
\maketitle
\begin{abstract}  
In line with the AI-native 6G vision, explainability and robustness are crucial for building trust and ensuring reliable performance in millimeter-wave (mmWave) systems. Efficient beam alignment is essential for initial access, but deep learning (DL) solutions face challenges, including high data collection overhead, hardware constraints, lack of explainability, and susceptibility to adversarial attacks. This paper proposes a robust and explainable DL-based beam alignment engine (BAE) for mmWave multiple-input multiple-output (MIMO) systems. The BAE uses received signal strength indicator (RSSI) measurements from wide beams to predict the best narrow beam, reducing the overhead of exhaustive beam sweeping.  To overcome the challenge of real-world data collection, this work leverages a site-specific digital twin (DT) to generate synthetic channel data closely resembling real-world environments. A model refinement via transfer learning is proposed to fine-tune the pre-trained model residing in the DT with minimal real-world data, effectively bridging mismatches between the digital replica and real-world environments.  
To reduce beam training overhead and enhance transparency, the framework uses deep Shapley additive explanations (SHAP) to rank input features by importance, prioritizing key spatial directions and minimizing beam sweeping. It also incorporates the Deep k-nearest neighbors (DkNN) algorithm, providing a credibility metric for detecting out-of-distribution inputs and ensuring robust, transparent decision-making. Experimental results show that the proposed framework reduces real-world data needs by 70\%, beam training overhead by 62\%, and improves outlier detection robustness by up to 8.5$\times$, achieving near-optimal spectral efficiency and transparent decision making compared to traditional softmax based
DL models.

\end{abstract}

\IEEEpeerreviewmaketitle

\section{Introduction}
\noindent  \IEEEPARstart{I}{nternational} Telecommunication Union’s IMT-2030 framework identifies ``integrated artificial intelligence (AI) and communication" as a key pillar for 6G networks, emphasizing the seamless embedding of AI into communication systems. 
The widespread adoption of AI techniques
in future 6G networks requires addressing critical challenges of \textit{explainability} and \textit{robustness} \cite{hamon2020robustness}. Explainability is vital for fostering trust in AI-driven systems, allowing network operators to understand, validate, and troubleshoot decisions made by deep learning (DL) models \cite{our_work}. Robustness ensures reliable performance by safeguarding against out-of-distribution inputs and adversarial attacks, which are common in dynamic and complex wireless environments. 

One of the key areas where these challenges emerge is beam alignment during initial access (IA) in wireless communications. In current 5G standards, beam alignment involves base stations (BS) sweeping beams using reference signals, while user equipment (UE) measures received signal strength indicators (RSSIs) and reports the strongest beam back to the BS \cite{ tutorialBA}. Although brute-force beam search guarantees optimal results by exhaustively evaluating all beam pairs, it incurs significant training overhead \cite{surveyBA0}.  Standard approaches often use quantized beams, which distribute energy across the angular space via codebooks, such as discrete Fourier transform (DFT) codebooks. While DFT codebooks ensure broad coverage, they often lack the granularity for precise beam alignment. To address this, oversampled DFT (O-DFT) codebooks provide finer granularity at the cost of increased beam training overhead, as more beams need to be evaluated during the alignment process. Generally, the beam sweeping time dominates the overall IA time, as it requires a considerable number of wide sensing beam measurements and reporting to cover the angular space. It is necessary to efficiently identify the optimal beam pairs between a BS and the UEs. 

While AI-aided techniques have demonstrated significant gains in beam alignment efficiency \cite{surveyBA1},  their opaque nature leads to a lack of transparency in decision-making and vulnerability to anomalies, which remain key obstacles. Existing literature often falls short in providing systematic methods to explain feature importance or align DL model behavior with wireless key performance indicators (KPIs). Moreover, training such models typically requires large-scale datasets, which are challenging to obtain due to hardware constraints and the high overhead associated with large antenna arrays. Training a DL-based beam alignment engine (BAE) typically requires large datasets, ranging from \(100\)K to \(1,000\)K samples \cite{FastIA, BA6}, which are costly to collect in deployed systems. Recent advancements in digital twins (DTs), demonstrated in \cite{alkhateeb0}, offer a promising solution to address these challenges by reducing or eliminating the dependence on extensive real-world channel data.  Addressing these aforementioned challenges is critical for developing DL-based beam alignment solutions that are efficient, transparent, and resilient while minimizing reliance on large-scale datasets. These advancements are vital for achieving standardization and facilitating the practical deployment of such technologies in next-generation communication systems.

\subsection{ Relevant Works}

\textbf{DL-based beam selection approaches}: Compared to traditional beam alignment methods, DL-based solutions have attracted great interest in reducing the overhead of beam training by providing environmentally adaptable, data-driven, and hardware-compatible solutions, wherein deep neural networks (DNNs) can extract features from additional
information such as auxiliary contextual information \cite{CI1}, \cite{morais2023position} and historical data by capturing temporal behavior of millimeter wave (mmWave) channels \cite{lim2021deep} and prior received signals \cite{ma2021deep}, enabling efficient beam prediction.  Deep reinforcement learning (DRL) and generative adversarial network (GAN)-based approaches have also been proposed to acquire efficient beam
management policies through iterative interactions with the
environment \cite{FL_DRL}, 
design of site-specific beam codebooks  \cite{OS_Asmaa}, or through dataset amplification via deep generative model,  bypassing the need for explicit channel state information (CSI) \cite{GANs-CE}.

\textbf{Explainable AI (XAI) based approaches}: Despite DL’s impressive performance in the beam alignment process \cite{CI1, morais2023position, lim2021deep, FastIA, BA6,ma2021deep, FL_DRL, OS_Asmaa, GANs-CE}, the underlying model lacks a systematic and explainable approach for input selection, along with mechanisms
that verify and explain the models’ decisions. This flip side of the highly performing yet complex DL
models, such as DNNs,  being ill-disposed
to direct interpretation restricts their adoption in mission-critical services. To strike a good balance between explainability and AI
model performance, XAI is
an emerging paradigm that aims to scrutinize the decisions made by opaque AI models \cite{our_work}, \cite{xai-oran}. 
Post-hoc XAI methods for feature selection have been used to identify key attributes that influence model decision-making and explain predictions of trained models \cite{XAI1, XAI5,  XAI4, khanTCOM, XAI_ICC}. Among the post-hoc XAI methods,  feature attribution techniques, such as Shapley Additive Explanations (SHAP) \cite{lundberg}, can highlight the features that most impact the model's predictions. Along this line of research, classical XAI methods and SHAP have been applied to identify and interpret wireless KPIs in 5G network slicing \cite{XAI1}, \cite{XAI5},  network intrusion detection  \cite{XAI4} and radio resource management \cite{khanTCOM}, \cite{XAI_ICC}. Additionally, SHAP-assisted composite reward mechanisms have been designed for DRL agents in 6G network slicing \cite{XAI_DRL}. The explainability offered by the aforementioned works \cite{XAI1, XAI5,   XAI4, XAI_ICC}  is coupled with the availability of large-scale wireless datasets without an established guiding principle or systematic methodology for selecting the inputs that significantly influence final decisions. We anticipate that XAI can play a crucial role in addressing this gap by offering methodologies for explainable input feature selection.

\textbf{Robustness against adversarial inputs}: DL-based beam classifiers are vulnerable to out-of-distribution inputs, which undermines resilience against adversarial attacks and reduces transparency—key requirements for standardization and commercial deployment \cite{our_work}.
DL frameworks often overestimate confidence in outlier inputs due to poor calibration of softmax output probabilities, leading to reduced beam prediction accuracy during inference. To address these issues, recent research leverages the modularity of neural networks, utilizing representations learned at each hidden layer. A nearest-neighbor search in these representation spaces can enhance robustness by evaluating the similarity of test inputs to the training data \cite{Deepknn}, \cite{khanexplainableICC25}. This approach provides certified defense mechanisms against adversarial examples and outliers, improving the reliability and security of DL-based beam classifiers.

\textbf{DT-assisted training}: 
DL-based beam alignment engine (BAE) should also be able to improve and adapt continuously (or at least periodically). This
is a major open challenge not well-addressed to date, as most
existing approaches assume extensive offline training prior to deployment, and when or how to re-train is also not well
understood. 
DTs offer a promising solution to reduce or eliminate the need for extensive real-world channel data by simulating wireless environments using high-fidelity 3D models and efficient ray tracing \cite{alkhateeb0}.  DTs use detailed environmental information, such as object positions, dynamics, shapes, and materials, to create accurate digital replicas. These replicas enable precise real-time simulations of wireless channels, supporting various communication tasks, including physical layer reliability \cite{reliableDT}, beam prediction \cite{alkhateeb1}, and channel compression and feedback \cite{alkhateeb2}. For beam prediction, DTs can simulate channels within digital replicas of the wireless environment and generate synthetic data closely mirroring real-world conditions. By fine-tuning the BAE with limited real-world data augmentation, site-specific DTs enable continuous model updates, effectively improving the model's generalization ability and robustness to variations in the input data. Classifiers trained on DT-generated data can be deployed rapidly across base stations and calibrated through few-shot transfer learning \cite{alkhateeb0, alkhateeb1}, ensuring models remain accurate and up-to-date while minimizing real-world data acquisition overhead. 

While DT-assisted solutions \cite{alkhateeb0, alkhateeb1,alkhateeb2}  offer notable efficiency improvements, they lack a focus on transparency and explainability, limiting their ability to foster trust in DL-based systems. 
Particularly, we highlight three pivotal challenges in existing beam prediction solutions \cite{surveyBA1, CI1, morais2023position, lim2021deep, FastIA, BA6,ma2021deep, FL_DRL, OS_Asmaa, GANs-CE, alkhateeb1} that must be effectively addressed:
\begin{enumerate}
    \item \textbf{Dependence on large-scale channel data}: DL-based approaches rely on received power measurements, which are influenced by channel quality. Obtaining extensive real-world channel datasets is challenging due to variations in channel, antenna, and network topology.
\item \textbf{Efficient beam selection}:  Prediction accuracy relies on both the number and selection of beams, as well as the need for site-specific beam probing to account for unique environmental characteristics. Identifying critical spatial directions tailored to specific deployment scenarios is essential for reducing beam sweeping and enhancing overall efficiency.
\item \textbf{Explainability and robustness}: DL models often lack transparency, making it difficult to quantify prediction confidence. Inaccurate predictions can degrade system performance, especially in real-world scenarios.
\end{enumerate}

\subsection{Contributions}
 {This paper presents the first work to provide an explainable and robust DL-based BAE for predicting mmWave beams during the IA process. A novel methodology integrating XAI is devised to select promising sensing beams with minimal beam sweeping overhead, while ensuring robustness in adversarial settings and offering insights into model failures, such as beam misalignment.
Main contributions are summarized as follows:
\begin{itemize}
    \item A DL-based beam alignment framework is proposed that leverages RSSI feedback from a finite set of sensing beams (DFT codebook) to predict optimal narrow beams from the O-DFT codebook for IA and data transmission. By employing a site-specific DT with accurate ray tracing, synthetic datasets resembling real-world environments are generated for model training. Further, an innovative transfer learning-based approach is devised for periodic refinement of the trained model using small amounts of real-world data augmentation, bridging the gap between synthetic and actual data distributions.

    \item  A novel two-stage systematic explainability framework leveraging feature relevance-oriented XAI is proposed  to identify the most influential sensing beams. The former stage involves generating feature importance ranking of the model's inputs using the SHAP-based importance scores. The latter stage exploits these rankings to reduce the feature input set to the DNN by removing the least important features from the model’s input. The proposed approach reduces beam sweeping overhead by at least $62\%$ while maintaining top-$k$ accuracy comparable to models trained on the full feature set. Moreover, compared to an exhaustive search over all O-DFT beams, the proposed DL-based approach reduces beam training overhead by $\approx 89\%$ while maintaining close to optimal spectral efficiency performance. 

    \item A  beam classifier integrating the deep k-nearest neighbors (DkNN) approach is proposed to enhance prediction explainability and robustness of the beam predictions against outlier inputs. The robust framework combines k-nearest neighbors with internal neural network representations, enabling reliable outlier detection and quantification of prediction credibility. This approach ensures robustness in adversarial settings and provides insights into model failures, such as beam misalignment. Simulations demonstrate that the DkNN-enhanced classifier improves outlier detection robustness by up to $8.5\times$ and provides superior interpretability compared to softmax-based classifiers.
\end{itemize}}

{The rest of the paper is organized as follows. Section \ref{sec:system} describes the mmWave communications system model and assumptions used in the paper. Section \ref{problem_outline} describes the problem formulation for the
beam alignment task and outlines the system operation of our proposed DT-aided explainable and robust beam alignment framework. Section \ref{sec:DL} presents the DL-based beam
alignment strategy, the DT-based data generation process, and the model refinement strategy for fine-tuning the pre-trained model. Section \ref{XAI} explains the proposed systematic XAI-based methodology for generating the feature importance ranking and describes the proposed feature selection algorithm. Section \ref{Robust-BAE} presents the proposed model robustness assessment framework for identifying outlier/adversarial inputs.  Section \ref{sec:simulation} evaluates the performance of the proposed solution strategy. Finally, conclusions and future research directions are presented in Section \ref{sec:conclusion}.}

\begin{figure}
    \centering
{\includegraphics[width=0.7 \linewidth, height=6.5cm]{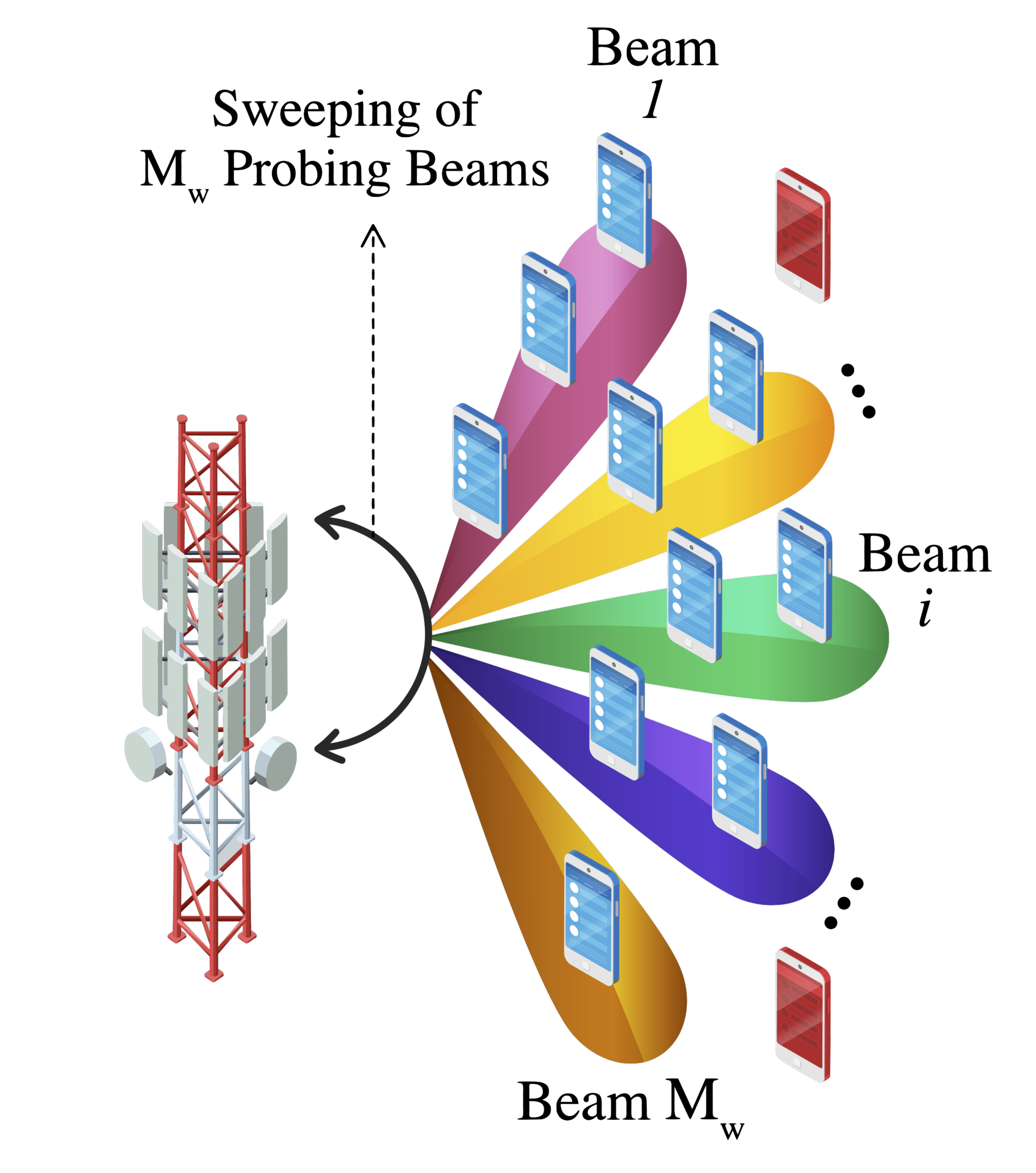}}
        \setlength{\abovecaptionskip}{-3pt}
    \setlength{\belowcaptionskip}{-15pt}
    \caption{System model illustration for the beam alignment in mmWave communication system.}
    \label{fig:sys_model}
\end{figure}

\begin{figure*}
    \centering
    \includegraphics[width=0.9\linewidth]{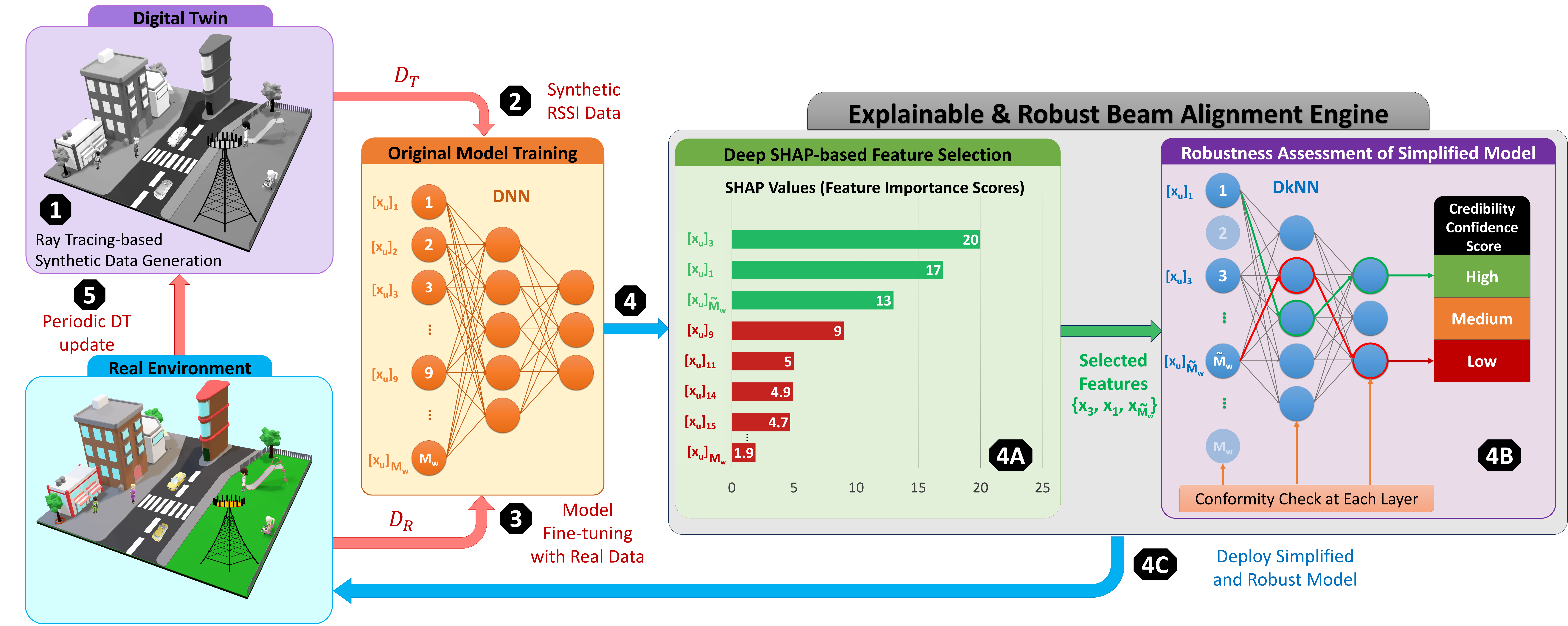}
    \setlength{\belowcaptionskip}{-10pt}
    \caption{Illustration of the proposed DT-aided explainable and robust beam alignment framework.}
    \label{proposed_flow}
\end{figure*}

\section{System Model} \label{sec:system}

{We consider a narrowband downlink mmWave communication system, where the BS features a uniform linear array (ULA) with $N_{\mathrm{BS}}$ antenna elements to communicate with $N_\mathrm{U}$ single-antenna
UEs}, as illustrated in Fig. \ref{fig:sys_model}.  We concentrate on the scenario of multi-user beamforming, where the BS communicates with every UE using only a single stream. Notably, for mmWave frequencies, channels often
exhibit sparsity in the angular domain, leading to a constrained
number of channel paths, $L$. The narrowband mmWave channel from the BS to the  $\mathrm{UE}_u$  can be described by a multipath geometric channel model \cite{heath2016overview} as

\begin{equation}
    \mathbf{h}_u =  \alpha_{u,l} \mathbf{b}\left(\phi_{u,l}\right),
\end{equation}
where  $\alpha_{u,l}$ represents the complex path gain for the $l$-th path of the $u^{th}$ UE, $\phi_{u,l} $ is the angle of departure for the $l$-th path, and $\mathbf{b}\left(\phi_{u,l}^{}\right)$  represents the beam steering vector, which is given by 
\begin{equation}
    \mathbf{b}(\phi_{u,l}) = \frac{1}{\sqrt{N_{\mathrm{BS}}}} \left[1, e^{j \frac{2 \pi}{\lambda} d \sin(\phi_{u,l})}, \dots, e^{j (N_{\mathrm{BS}} - 1) \frac{2 \pi}{\lambda} d \sin(\phi_{u,l})}\right]^T,
\end{equation}
where $\lambda$  is the signal wavelength given by $\lambda = \frac{c_0}{f_0}$, $c_0$ is the speed of light, $f_0$ is the carrier frequency, and $d=\frac{\lambda}{2}$ denotes the antenna spacing. {Note that for the narrowband channel model, we made the implicit assumption (very common in most beamforming and array processing literature) that the
spatial directions are frequency-independent and the communication bandwidth $B$ is much smaller than the carrier frequency $f_0$, such that the array response is essentially constant with $f \in [f_0 - B/2, f_0 + B/2]$.}\footnote{ {For wideband mmWave system,  the required phase shifts are frequency-dependent. The result is that beams for frequencies other than the carrier “squint”
as a function of frequency \cite{wideband_squint}. }
} To mitigate the hardware cost and power consumption of a fully digital system, we adopt analog-only beamforming where the BS has a single common transmit/receive radio frequency (RF) chain shared by $N_{\mathrm{BS}}$ antennas. Hence, the beamforming vector at the BS is given by
\begin{equation}
\mathbf{w}=\frac{1}{\sqrt{N_{\mathrm{BS}}}}\left[e^{j \varphi_1}, \ldots, e^{j \varphi_i}, \ldots, e^{j \varphi_{N_{\mathrm{BS}}}}\right]^{\top} \in \mathbb{C}^{N_{\mathrm{BS}} \times 1},
\end{equation}
where $\varphi_i$ is the phase shift of $i$-th antenna element.
We assume that the BS adopt a beamforming codebook $\mathbf{W}=$ $\left\{\mathrm{\mathbf{w}}_1, \ldots, \mathrm{\mathbf{w}}_Q\right\} $ incorporating $Q$ pre-defined beamforming vectors.  The vectors in $\mathbf{W}$ satisfy $\left\|\mathrm{\mathbf{w}}_{q}\right\|^2=1, \forall q \in\{1, \ldots, Q\}$  to accommodate the constant-modulus constraint of analog beamforming architecture. 

During the beam sweeping process, the BS periodically
transmits symbols  $s_w \in \mathbb{C}$ to the UEs through the beams defined by the matrix $\mathbf{W} \in \mathbb{C}^{N_{\mathrm{BS}} \times Q} $. Following the beam sweeping process, the complex received signal at the $\mathrm{UE}_u$, using the $q_u$-th beamforming vector $\mathbf{w}_{q_u}$, can be expressed as

\begin{equation}
   {r}_{u} = \sqrt{P_{\mathrm{BS}}} \mathbf{h}_{u}^H \mathbf{w}_{q_u} s_{w} + {z}_{u},
\end{equation}
where $P_{\mathrm{BS}}$ denotes the BS transmit power, $\mathbf{h}_u \in \mathbb{C}^{N_{\mathrm{BS}} \times 1}$ is the channel vector, and ${z}_{u}$ represents the additive complex noise with power $\sigma_z^2$. Then, with unit-power transmitted symbols, the signal-to-noise ratio (SNR) at  the $\mathrm{UE}_u$ can be expressed as
\begin{equation}
    \text{SNR}_u = \frac{P_{\mathrm{BS}} \left| \mathbf{h}_{u}^H \mathbf{w}_{q_u} \right|^2}{\sigma_z^2}.
\end{equation}

\section{Problem Formulation and Solution Methodology}\label{problem_outline}
This section describes the problem formulation for the beam alignment task and introduces our proposed DT-aided explainable and robust beam alignment framework. 
\subsection{Problem Formulation}

{
For a given BS-UE pair, the optimal beam index $ q_{u}^*$ can be identified by selecting the beam that maximizes the SNR:
\begin{align} \label{beam_selection}
    q_{u}^* &= \underset{q_u \in \left\{1,2,\dots,Q\right\}}{\arg \max} 
    \left( \frac{P_{\mathrm{BS}}\left| \mathbf{h}_{u}^H \mathbf{w}_{q_u} \right|^2 }{\sigma_z^2} \right) \notag \\
    &= \underset{q_{u{}} \in \left\{1,2,\dots,Q\right\}}{\arg \max} 
    \left( \left| \mathbf{h}_{u}^H \mathbf{w}_{q_u} \right|^2 \right).
\end{align}
}

Notably, the overall number of beams swept remains constant,
unaffected by the number of UEs, as multiple UEs can
measure the sensing beams concurrently, utilizing shared time
and frequency resources. It is worth noting that identifying $q_{u\mathbf{}}^*$ through exhaustive beam search over $Q$ codewords results in significant beam sweeping overhead. To address this issue, we leverage the exploration capabilities of DL to identify optimal
beam directions tailored to specific sites.

\subsection{Solution Methodology} \label{sec:prob-sol}
To overcome the challenges with existing DL-based beam alignment solutions in terms of lacking real training datasets, high beam training overhead, and lack of transparency in decision-making, we propose a framework where the BS uses the digital replica of the physical environment to enhance network performance and provide an explainable and robust solution for the beam alignment task. As illustrated in Fig. \ref{proposed_flow} and introduced in the sequel sections, the system operation proceeds as follows:
\begin{itemize}
    \item[\large \textcircled{\small{1}}] Synthetic channel data is first generated in a site-specific DT using ray tracing, [c.f. Section \ref{data_generation_sub}].    
    \item[\large \textcircled{\small{2}}] The synthetic datasets are then utilized to train the DL-based BAE model, [c.f. Section \ref{data_generation_sub}]. 
    \item[\large \textcircled{\small{3}}] The DL-based BAE, initially trained on synthetic data, is then fine-tuned by augmenting it with a small portion of real-world data. This enhances the generalization capability of the DT models and mitigates the distribution gap between real and synthetic data, [c.f. Section \ref{tl_sub}].
    \item[\large \textcircled{\small{4}}] The post-hoc explainability and robustness assessment then begins with \textcircled{\footnotesize{A}}, where the deep SHAP method from XAI is employed to assign importance scores to the input features. This enables the selection of a subset of features, such as RSSIs, that significantly contribute to the DL model's predictions [c.f. Section \ref{XAI}]. In \textcircled{\footnotesize{B}}, the robustness of the model trained on this selected subset is evaluated. This process involves inspecting the internal hidden layers using a nearest-neighbor search to calculate conformity scores, which measure the similarity of test inputs to the training data. This step helps identify adversarial and outlier data, ensuring the model's reliability [c.f. Section \ref{Robust-BAE}]. Finally, \textcircled{\footnotesize{C}}, the refined model is deployed in a real-world environment for online operation.
    \item [\large \textcircled{\small{5}}] The DT is periodically updated where the BS collects real data through user feedback during deployment to compensate for impairments in the DT data and synchronize the DL model, e.g., when the environment experiences significant changes, by repeating \textcircled{\small{1}}, \textcircled{\small{2}}, \textcircled{\small{3}}, and \textcircled{\small{4}}. 
\end{itemize}

\section{Deep learning framework for beam alignment} \label{sec:DL} 
This section describes the details of the DL-based beam alignment strategy and the DT-based data acquisition. We first describe the beam selection problem as a nonlinear mapping from the collected RSSIs values to the optimal narrow beam index. Then, we detail the  DT-assisted data acquisition approach, followed by the transfer learning-based model refinement strategy to overcome the modeling mismatches between the physical environment and its digital replica.

\subsection{DL-based Beam Alignment}
{The proposed DL-aided BAE system falls into the
beam sweeping framework  compatible with the 5G NR standard. The proposed solution leverages the RSSI values over a small set of compact wide sensing beams $M_{\mathrm{w}}$ (i.e., beams from DFT codebook with $M_{\mathrm{w}}\ll Q$ ), to select the best narrow beam from the O-DFT codebook of size $Q$, thus, avoiding exhaustive extensive  search over the O-DFT codebook.} For beam-sweeping, the BS transmits the pilot signals over a smaller set of beams, each at a separate time slot.  All UEs connected to the BS measure and report the RSSI values of the $M_{\mathrm{w}}$ beams. It is assumed that beam sweeping, measurement, and reporting occur within the coherence time during which the channel remains constant \footnote{ {For low mobility applications, channel remains coherent sufficiently long to accommodate the full beam sweeping and reporting
process without significant channel variations. For high
mobility applications, the relevant
metric is the beam coherence time and is  generally much higher than the channel coherence time \cite{alkhateeb2018deep}.} }. The reported beam sweeping results for  UE  $u$ can be written as
\begin{equation}\label{rp}
\mathbf{x}_u={\left[[x_u]_1,\ldots,[x_u]_{M_\mathrm{w}} \right]}=\left[\left|\left[r_{u}\right]_1\right|^2 \cdots\left|\left[r_{u}\right]_{M_{\mathrm{w}}}\right|^2\right]^T ,
\end{equation}
where $\left[r_{u}\right]_i=\sqrt{P_{\mathrm{BS}}} \mathbf{h}_u^H \mathbf{w}_is_w+z_i$ is the received signal using the $i$-th sensing beam, $\forall i \in M_{\mathrm{w}}$.

In the proposed DL-based approach, a DNN is trained to  associate
the correct/best-oriented beam between the BS and
UE with the RSSIs over wide sensing beams as the DNN's inputs and the index of
the corresponding optimal narrow beam as the output. This beam selection problem is formulated as a classification task, where each classified category
corresponds to one candidate narrow beam. Particularly, the reported beam sweeping results from (\ref{rp}) are fed as input to DNN-based beam classifier denoted by $f\left(\cdot \hspace{0.05cm}; {\boldsymbol{\theta}}\right): \mathcal{X} \rightarrow \mathbb{R}^Q$, where $\mathcal{X} \in  \mathbb{R}^{M_\mathrm{w}} $ is the set of DNN's input corresponding to the RSSI values over $M_{\mathrm{w}}$ beams and $\boldsymbol{\theta}$ denotes the DNN's weight parameters vector.

One challenge with the DL approach is the demand for real-world training data. The DL model typically requires a dataset that captures the environmental characteristics. For the beam alignment task in  \eqref{beam_selection}, the goal is to assign the optimal beam index by identifying the spatial directions that provide the strongest RSSI values, {which are a function of channel characteristics $\mathbf{h}_u $, which in turn depend on the real environment  ($\mathcal{E}_u$) for the $\mathrm{UE}$ $u$, i.e., $\mathbf{h}_u = g(\mathcal{E}_u)$}, where $g(\cdot)$ represents the wireless propagation model \cite{alkhateeb0}. 
However, gathering real-world channel measurements is often a challenging and resource-intensive process that constitutes the main obstacle to deploying DL-based BAE in practice.

As a solution to this problem, transfer learning techniques can be used to transfer parts of information
in a network previously trained with a large dataset to another
network for which only a small training dataset is available \cite{survey_TL}. This process, known as domain adaptation, enables information to be shared across different environments. Assuming a large dataset
of synthetic measurements can be collected from a digital environment, which we refer to as
source domain (DT scenario), the DNN weights can be trained by a standard backpropagation algorithm. On the other hand, in a second environment, which we refer to as the destination
domain (real scenario), only a limited-size dataset is available since the destination domain corresponds to real-world deployment. Then, the information learned from
the source domain can be transferred to the destination via transfer learning \cite{deep-tl}, allowing the model to learn from previously encountered data and new data samples. Next, we detail the procedure for channel data acquisition and synthetic dataset construction by leveraging the site-specific DT to approximate the communication environment.


 {
}

\subsection{Site-specific DT-aided Data Acquisition  }\label{data_generation_sub}
 As illustrated in Fig. \ref{proposed_flow}, site-specific DT resides at the BS to generate synthetic channel data via ray tracing for learning and optimizing beam selection. This twin can model the communication environment $\mathcal{E}_u$ and the wireless propagation model $g(\cdot)$ through an electromagnetic (EM) 3D model ${\mathcal{E}_{u,\mathrm{T}} }$ and a ray tracing based approximation $g_\mathrm{T}(\cdot)$ of the wireless propagation model \cite{alkhateeb0}. The EM 3D model provides estimated data on the positions, shapes, and materials of the BS, UE, and surrounding objects, whereas ray tracing generates propagation paths using this data to create spatially consistent channels in the digital replica. The generated synthetic channel data from the DT mirrors the distribution of real-world data. In particular, the environment DT constructs a synthetic labeled dataset 
${\mathcal{D}}_\mathrm{T} = \left\{ \left(\mathbf{X}_\mathrm{T}, \mathbf{q}_\mathrm{T}\right) = {(\mathbf{x}}_{u,d}, {q}_{u,d}^{\star}) : d = 1, \cdots, {{D}}_\mathrm{T}
 \right\}$, comprising ${D}_\mathrm{T}$ samples, where ${\mathbf{x}}_{u,d}$ represents the RSSI values as input features for the 
$d$-th sample, ${q}_{u,d}^{\star}$ is the O-DFT beam index with the highest RSSI as the target label, generated using \eqref{beam_selection_DT}. Substituting the 3D model $\mathcal{E}_{u,\mathrm{T}}$ and the ray tracing approximation of channel $\mathbf{h}_{u,\mathrm{T}}={g}_{\mathrm{T}}(\mathcal{E}_{u,\mathrm{T}})$ into (\ref{beam_selection}), {the beam prediction task performed in the digital replica can be expressed as

\begin{equation} \label{beam_selection_DT}
{q}^{\star}_u=\underset{q_{u} \in\{1, \ldots, Q\}}{\arg \max }\left|\mathbf{w}_{q_u}^H \mathbf{h}_{u,\mathrm{T}}\right|^2.
\end{equation}
}

Furthermore, cross-entropy loss, which quantifies the dissimilarity between predicted probabilities and true labels, is applied to train the DNN's weight parameters, using the one-hot encoding of the optimal beam index 
in \eqref{beam_selection} as the ground truth label. {To  facilitate the loss function design during training, we  obtain the
ground-truth label by searching the optimal beam direction for each sample  in the 0-DFT codebook by performing noise-free exhaustive search.}
The  loss function is formulated as: 
\begin{equation} \label{ref_loss}
\mathcal{L}(\boldsymbol{\theta} , \mathcal{D}_T) = - \frac{1}{D_\mathrm{T}} \sum_{d=1}^{D_\mathrm{T}} \sum_{q_u=1}^{Q} p_{u,d,q_u}^\star \log \widehat{p}_{u,d,q_u},
\end{equation} 
{where $p_{u,d, q_u}^\star$ is the true distribution over  labels indicating that the $d$-th data sample belongs to $q_u$-th class, such that $p_{u,d, q_u}^\star = 1$ if the $q_u$-th candidate narrow beam is the actual
optimal beam ($q_u^\star$) and $p_{u,d, q_u}^\star = 0$ otherwise, and $\widehat{p}_{u,d, q_u} \in$ $\mathbb{R} \rightarrow (0,1)$ represents the predicted probability distribution, obtained by applying a softmax function to the logits produced by the beam classifier. 
}

 In practice, constructing a DT that perfectly mirrors the real-world environment is inherently challenging as the DT model approximating the actual environment is prone to distribution shifts and inaccuracies in parameter estimations, often leading to degraded beam alignment accuracy. These inaccuracies arise from the inherent modeling impairments of the DT, which limit the performance of models trained exclusively on synthetic data. As a remedy, a small amount of real-world data augmentation is utilized to fine-tune the pre-trained model (i.e., transfer learning) to transcend the digital replica.

\subsection{Transfer Learning for Model Refinement}\label{tl_sub}

As illustrated in Fig. \ref{proposed_flow}, the DT keeps a dynamically updated representation of the actual environment. Copies of the pre-trained DNN are further refined by re-training the pre-trained model using a small amount of real-world data augmentation to achieve similar performance as the model
solely trained on real-world data. In particular, transfer learning is leveraged to adapt the model previously trained on the large synthetic dataset ${\mathcal{D}}_\mathrm{T}$ and retrained by augmenting training samples 
 from a real-world dataset.  The real-world data,
for instance, can be collected from the feedback provided by the
UEs during the online operations.  

While the training samples in  ${\mathcal{D}}_\mathrm{T}$  can be continuously
generated by the DT in the background, the BS keeps collecting the RSSI values as feedback from the users to construct the real dataset $\mathcal{D}_{\mathrm{R}}=\left\{(\mathbf{X}_\mathrm{R},\mathbf{q}_\mathrm{R})\right\} = {(\mathbf{x}}_{u,d}, {q}_{u,d}^{\star}) : d = 1, \cdots, {{D}}_\mathrm{R}\}$, defined similarly as ${\mathcal{D}}_\mathrm{T}$ with size ${D}_\mathrm{R}$. 
Then, by augmenting a small proportion, e.g., $20-30\%$ of real data samples from  $\mathcal{D}_{\mathrm{R}}$, the DL-based BAE is refined to achieve comparable performance to the model purely trained on
real-world data $\mathcal{D}_{\mathrm{R}}$. Therefore, the model refinement process aims to minimize the loss function in \eqref{ref_loss} on the augmented dataset $\mathcal{D} = \mathcal{D}_{\mathrm{R}} \cup {\mathcal{D}_\mathrm{T}}=\left\{( \mathbf{X},\mathbf{q})\right\} = {(\mathbf{x}}_{u,d}, {q}_{u,d}^{\star}) : d = 1, \cdots, {{D}}\}$ comprising $D$ data samples instead of $D_{\mathrm{T}}$ samples.


\section {XAI Guided  Beam Training Reduction}\label{XAI}
The DL-based beam alignment strategy described in the previous section relies on sweeping sensing beams at the BS to collect RSSI values \eqref{rp}, which are used as input features to the DL model.  To reduce sweeping overhead and improve alignment efficiency, we propose an XAI-based feature selection approach to identify a subset of input RSSIs. This approach leverages the fact that certain RSSIs contribute more significantly to the beam selection process, enabling a reduction in beam training overhead by avoiding sweeping the entire set of sensing beams. This section introduces an XAI-based method for optimizing sensing beam selection by analyzing the contributions of RSSI values across candidate beams. Using SHAP \cite{Deepshap}, a robust framework for computing local and global feature importance in machine learning models, we assign importance scores to the inputs of a pre-trained model. These scores are then utilized to derive an efficient feature selection strategy, minimizing the \textit{beam sweep time} required during IA.
It is worth mentioning that the feature selection process is conducted {offline} and in a {post-hoc} manner, meaning that the pre-trained model trained on the augmented dataset $\mathcal{D}$  is readily available for SHAP-based analysis.

As illustrated in Fig. \ref{proposed_flow}, SHAP is used for the selection of a subset of RSSI inputs collected  over ${M}_{\mathrm{w}}$ sensing beams facilitating an understanding of feature significance. {Instead of  measuring the RSSI values $\mathbf{x} = \{x_1, \ldots, x_{\mathrm{M_w}}\}$, where each $x_i$ corresponds to the RSSI measurements from the $i$-th sensing beam, a subset of these RSSI measurement is used to represent the subset of features, i.e, the RSSI values, deemed most important for the model's prediction. We denote the size of SHAP-based selected input features as $ \widetilde{M}_{\mathrm{w}}$ such that $\widetilde{M}_{\mathrm{w}} \leq  M_{\mathrm{w}}$.} Specifically, we aim to identify the most influential features for beam prediction to maximize accuracy while minimizing the number of sensing beams used. {For \ $\widetilde{M}_{\mathrm{w}}=32$, cycling through all possible combinations  would require checking for \( 2^{32}= 4.29 \times 10^9 \) subset of input features to identify the one that performs the best, rendering exhaustive evaluation impractical}. For simplicity, we drop the index $u$ in subsequent notations. In the following, we briefly introduce Shapley values, describe SHAP, and detail the proposed XAI-based feature selection algorithm for sensing beam optimization.

\subsection{ Preliminaries on  Shapley Values and SHAP}
Shapley values quantify the marginal contribution of each input feature \(x_i\), \(\forall i \in \{1, \dots, M_{\mathrm{w}}\}\), to a model's prediction \(f(\mathbf{x})\), where \(x_i\) is the ${i}^{\text{th}}$ element of input  \(\mathbf{x}\) \eqref{rp}. The Shapley value for feature \(x_i\) is computed by evaluating all possible coalitions of features and averaging the weighted differences in predictions with and without \(x_i\). For a prediction model \(f(\cdot)\), the Shapley value \(\psi_i( \mathbf{x})\) is defined as:

\begin{equation}\label{shapley_equation}
\small
\psi_i(f, \mathbf{x}) = \sum_{\mathbf{z'} \subseteq \mathbf{x'}} \frac{|\mathbf{z'}|!(M_{\mathrm{w}} - |\mathbf{z'}| - 1)!}{M_{\mathrm{w}}!} \left[ f(\mathbf{z'}) - f(\mathbf{z'} \setminus \{x_i\}) \right],
\end{equation}\normalsize
where \( \psi_i(f,\mathbf{x}) \) represents the Shapley value for feature \( x_i \), quantifying its contribution to the model's output, \(M_{\mathrm{w}}\) is the total number of features, \(\mathbf{z'}\) is a subset of features, \( |\mathbf{z'}| \) denotes the number of original features included in the subset \( \mathbf{z'} \), \(\mathbf{x'}\) is the set of all original features, \(f(\mathbf{z'})\) is the prediction on \(\mathbf{z'}\), and \(f(\mathbf{z'} \setminus \{x_i\})\) is the prediction without \(x_i\). The term \( \frac{|\mathbf{z'}|!(M_{\mathrm{w}} - |\mathbf{z'}| - 1)!}{M_{\mathrm{w}}!} \) is the Shapley weight, which ensures a fair attribution by averaging over all possible feature permutations. The Shapley value \(\psi_i(f, \mathbf{x})\) measures \(x_i\)'s contribution to the transition from the model's expected output (without \(x_i\)) to the actual prediction. These values include both magnitude and sign, indicating the significance and direction of each feature's impact.

Exact computation of Shapley values requires evaluating all \(2^{M_{\mathrm{w}}}\) feature combinations, which is computationally infeasible for DL models with a large number of inputs. To address this, {Deep-SHAP} offers an efficient approach by combining Shapley values with {DeepLIFT}, a method that decomposes the output prediction of a neural network by backpropagating contributions of all neurons in the network to the input features \cite{Deepshap, lift}. Deep-SHAP leverages DeepLIFT's per-node attribution property to aggregate Shapley values computed for smaller network components into values for the entire DNN. Deep-SHAP estimates feature importance by calculating the difference between input features and their corresponding reference values, which are derived from a \textit{background} dataset (a subset of the training data). These reference values are propagated through the DNN to precompute reference input and output values for each neuron. When explaining a specific prediction, each input feature is assigned a score that quantifies how its deviation from the reference contributes to the difference in the output. By integrating over multiple samples from the background dataset, Deep-SHAP efficiently approximates Shapley values without the exponential computational cost of evaluating all feature combinations \cite{lift}. This allows for scalable and interpretable explanations of DL model predictions.\footnote{For more details, we refer interested readers to \cite{Deepshap, lift}.}

\subsection{Deep-SHAP-based Feature Selection Methodology}
Deep-SHAP based feature importance analysis requires three inputs: a pre-trained DL model, a background dataset, and a set of data points to be explained. In our analysis, a proportion of the dataset used to train the DL model is used as the \textit{background } dataset $\mathcal{D}_\mathrm{BG}$  
 and the data samples to be explained come from the \textit{test} dataset $\mathcal{D}_\mathrm{test}$. 
 As described previously, the dataset $\mathcal{D}_\mathrm{BG}$ is used to compute baseline predictions, against which the contributions of individual features in the test dataset are measured \cite{Deepshap}.
Our proposed methodology consists of two stages: (1) generating feature importance rankings using Deep-SHAP and (2) selecting a reduced subset of input features based on the generated rankings.

 In the first stage, the Deep-SHAP explainer from the SHAP library \cite{lundberg} is instantiated using the pre-trained model \(f(\mathbf{x}; \boldsymbol{\theta})\) and dataset \(\mathcal{D}_\mathrm{BG}\), which represents the baseline reference values for feature importance calculations. {For each test sample in \(\mathcal{D}_\mathrm{test}\), the explainer evaluates the contribution of individual features to the model's prediction and outputs the SHAP values signifying the features importance scores. }

Formally, Deep-SHAP produces local explanations as importance scores in the form of SHAP values $\boldsymbol{\psi}(\mathbf{x}) = \left[\psi^d_{1,1}, \ldots, \psi^d_{i,q}, \ldots, \psi^d_{M_\mathrm{w}, Q}\right],$ where \(\psi^d_{i,q}\) denotes the impact of feature \(x_i\) on the observed output $q$ for data sample $d$,  quantifying the contribution of feature $x_i$ to $q$-th beam class label.   Here, \(i \in \{1, \ldots, M_\mathrm{w}\}\) indexes the features, \(q \in \{1, \ldots, Q\}\) indexes the beam labels, and \(d\in \{1, \ldots, D_\mathrm{test}\}\) indexes the data samples to be explained. Notice that Deep-SHAP provides SHAP values for each sample instance $d$ and each probable beam index (output class label). Global explanations are obtained by averaging the absolute SHAP values across all data samples and beam labels:
\begin{equation}
    \bar{\psi}_i = \frac{1}{|D_\mathrm{test}| \cdot Q} \sum_{d=1}^{|D_\mathrm{test}|} \sum_{q=1}^Q |\psi^d_{i,q}|, \label{equ:d-SHap}
\end{equation}
resulting in a vector of mean absolute SHAP values \(\bar{\boldsymbol{\psi}}(\mathbf{x}) = \left[\bar{\psi}_1, \ldots, \bar{\psi}_i, \ldots, \bar{\psi}_{M_\mathrm{w}}\right]\). Sorting this vector in descending order provides the global feature/beam importance rankings, where the first position represents the most important feature. These rankings guide feature selection and can be visualized to provide network operators with interpretable explanations of influential beam patterns.

\begin{algorithm}[!t]\small
  \SetAlgoLined
  \caption{Deep-SHAP based Feature Selection}
  \label{algo_fixed}
   \KwIn{  $f(\mathbf{x}; \boldsymbol{\theta})$ , $\mathcal{D}_\mathrm{BG}, \delta$} 
  \KwOut{Subset of input features $\mathbf{x}_S$ }
  
    \SetKwProg{myproc}{Stage 1: Beam importance rankings}{}{}
  \myproc{}{
  Instantiate the Deep-SHAP explainer\;
    Compute $\boldsymbol{\psi}(\mathbf{x})= \left[\psi^d_{1,1}, \ldots, \psi^d_{\mathrm{M_w}, Q}\right]$\; 
    Aggregate SHAP values as per \eqref{equ:d-SHap};\\
    $
     \bar{\boldsymbol{\psi}}(\mathbf{x}) \gets \text{argsort}\left( \left[\bar{\psi}_1, \ldots, \bar{\psi}_{\mathrm{M_w}}\right], \text{descending}\right)
    $\\
  }
  \SetKwProg{myproc}{Stage 2:  Threshold-based Feature Selection}{}{}
  \myproc{}{ 
      Compute $\mathcal{M}_{\mathrm{SHAP}}(\delta)$ using (\ref{shap_algo_eq})  \;
  }
  \KwRet{Subset of input features $ \mathbf{x}_S$}
\end{algorithm}

{ {In the second stage, we devise a feature selection strategy based on the relative significance 
of SHAP values, i.e., selecting all input features whose importance scores lie above a threshold \(\delta\), such that the sum of the SHAP values for the selected features accounts for a  fraction $\delta$ of the aggregated SHAP values.} Mathematically, the selection criterion for feature $x_i$ can be expressed as:

\begin{equation}\label{shap_algo_eq}
\mathcal{M}_{\mathrm{SHAP}}(\delta) = \left\{i \mid \sum_{i \in \mathbf{x}_S} \bar{\psi}_i \geq \delta \cdot \sum_{j \in \mathbf{x}} \bar{\psi}_j,\right\},
\end{equation}
where $\mathbf{x}_S$ represents the subset of important features in descending order of SHAP values., i.e, the RSSI inputs corresponding to the highest SHAP values, $\mathbf{x}= \{x_1, \ldots, x_\mathrm{M_w}\}$ represents the set of all features, $\delta$ is the threshold percentage. Here, \(\mathcal{M}_{\mathrm{SHAP}}(\delta)\) is the index set of selected features satisfying the threshold criterion. Using this index set, the subset of features \(\mathbf{x}_S = \{x_i \mid i \in \mathcal{M}_{\mathrm{SHAP}}(\delta)\}\) is formed, containing $\widetilde{M}_{\mathrm{w}}$ features, where $\widetilde{M}_{\mathrm{w}} = |\mathbf{x}_S|$. Since the SHAP values are typically concentrated among the top-ranked features (following a diminishing contribution trend), moderately high values of $\delta$ (e.g., $>70\%$), ensure that the most influential features are retained. 

The Deep-SHAP based feature selection process is summarized in Algorithm \ref{algo_fixed}. Algorithm \ref{algo_fixed} provides a systematic method to identify the optimal subset of RSSIs associated with the \( \widetilde{M}_{\mathrm{w}} \) sensing beams. The inputs of the algorithm are the pre-trained model to be explained $f(\mathbf{x}; \boldsymbol{\theta})$, the background dataset $\mathcal{D}_\mathrm{BG}$, 
 and a feature selection threshold $\delta$. {The  output is a subset of RSSI measurements $\mathbf{x}_S= \{x_1, \ldots, x_\mathrm{\widetilde{M}_w}\}$ selected according to the SHAP importance criterion in \eqref{shap_algo_eq}. A simplified model is then retrained in the DT using this subset of critical features $\mathbf{x}_S$, reducing model complexity (e.g., via feature selection) and beam sweeping time by minimizing the number of sensing beams to \( \widetilde{M}_{\mathrm{w}} \) instead of \( {M}_{\mathrm{w}} \).} 
 Next, this simplified model is used to assess the robustness of the beam classifier against adverserial inputs.

\section{Proposed Explainable and Robust Beam Classifier Framework}\label{Robust-BAE}
{ In this section, we devise a robustness assessment framework for the simplified DL-based beam classifier model proposed in Section \ref{XAI}. {Although the proposed Deep-SHAP-based feature selection scheme effectively reduces beam sweeping overhead by minimizing the input size and enhances robustness to noise by sweeping additional top-k predicted narrow beams from the 0-DFT codebook, the underlying DNN still lacks robustness to out-of-distribution/adversarial inputs. For instance, a jammer can sabotage the legitimate BS-UE communication system, by sending directional jamming signals, introducing outlier RSSI inputs that mislead the DL-based classifier and result in incorrect beam selection. Consequently, the system lacks robustness and explainability under adversarial conditions.}


{As shown in Fig. \ref{proposed_flow}, the pre-trained simplified DNN, denoted by 
$f^{'}\left(\mathbf{x}_S; {\boldsymbol{\theta}}\right): \mathcal{X}_S \rightarrow \mathbb{R}^Q$,  operates on a reduced input set  $\mathcal{X}_S \in \mathbb{R}^{|D| \times \widetilde{M}_\mathrm{w}}$.  This input set is obtained by selecting $\widetilde{M}_\mathrm{w}$ critical features from the original augmented dataset $\mathcal{D} \in \mathbb{R}^{|D| \times M_\mathrm{w}}$, 
as per the selection process defined in \eqref{shap_algo_eq}.} The proposed robust beam classifier adapts the deep k-nearest
neighbors (DkNN) approach from \cite{Deepknn} to analyze the simplified DNN's internal representations during testing by identifying inconsistent predictions with training data. The DkNN algorithm is model-agnostic and applicable to any pre-trained DL model with manageable computational complexity.}


{The simplified pre-trained DNN $f^{'}\left(\mathbf{x}_S; {\boldsymbol{\theta}}\right)$ is composed of  $L$ layers, where each layer is indexed by $\eta$, with  $\eta \in \{0, \dots, L-1\}$. } The DkNN  algorithm records the output of each layer $ f_\eta $ for every training point, thus obtaining a per-layer representation of the training data points paired with their respective labels. This per-layer representation is used to build a nearest neighbor classifier in the space defined by each layer to create a representation of the training data at each layer. { To efficiently identify nearest neighbors in the high-dimensional spaces produced by these layers, we use locality-sensitive hashing (LSH) \cite{andoni2015practical}, which finds similar representations based on cosine similarity. Given a test input, locality-sensitive hashing functions are first used to establish a list of candidate nearest neighbors: these are points that collide (i.e., are similar) with the test input. Then, the nearest neighbors can be found among this set of candidate points.  At every layer of the trained DNN, we use LSH to efficiently retrieve a set $\Pi$ of the $\widetilde{k}$ training samples whose representations are most similar to $f_\eta(\hat{\mathbf{x}}_S)$ based on cosine similarity.  Because the   $\widetilde{k}$ set of neighbors are from the training data, we have labels for each of these data points. The labels associated with these nearest neighbors in  $\Pi$ are collected for each layer, and these multi-sets of labels are used to compute the final beam selection prediction through a non-conformity check.}

For a given test input $\hat{\mathbf{x}}_S$, the non-conformity of a prediction, which is defined as the number of nearest neighboring representations found  in training data whose label is different from the candidate label $j \in \left\{1,2,\dots,Q\right\}$, is  expressed as: 
\begin{equation}\label{non_conformity}
\varrho(\hat{\mathbf{x}}_S, j)=\sum_{\eta \in {0, \dots, L-1} }\left|\left\{i \in \Omega_\eta: i \neq j\right\}\right|,
\end{equation}
where $\Omega_\eta$ is the multi-set of labels for the training points whose representations are closest to the test input's at layer $\eta$, and the operator $\left|\cdot\right|$ denotes the cardinality of a set, which is the number of elements contained within it. 

Before making inference decisions, i.e, beam predictions, we compute the nonconformity of a labeled holdout dataset $ ({X}^c, {Q}^c)$, sampled from the same distribution as the training data $ ({X}, {Q}) \in \mathcal{D}$ but not used for training. The holdout dataset provides an unbiased reference for nonconformity scores, ensuring statistically sound predictions by formalizing prediction validity through hypothesis testing: the larger the non-conformity, the weaker the training
data supports the prediction. The nonconformity values computed on the set of holdout data are defined as $\mathcal{C}$ =$ \left\{\varrho(\hat{\mathbf{x}}^c_u, {q}^c) : (\hat{\mathbf{x}}^c_u, q^c) \in ({X}^c, {Q}^c)\right\}$, and are then compared to the test input’s nonconformity score $\varrho(\hat{\mathbf{x}}_S, j)$ for each candidate beam label $ j $. For input $\hat{\mathbf{x}}_S$ with label $ j $, we calculate the empirical $p$-value, which represents the fraction of holdout nonconformity scores larger than the test input’s score,  as: 
\begin{equation}
p_j(\hat{\mathbf{x}}_S) = \frac{|\{\varrho \in \mathcal{C} : \varrho \geq \varrho(\hat{\mathbf{x}}_S, j)\}|}{|\mathcal{C}|}.
\end{equation}

The empirical $p$-value transforms nonconformity scores into a probabilistic framework, enabling interpretable predictions with quantifiable uncertainty (confidence) and support (credibility)  \cite{Deepknn}. A low $p$-value indicates that the test input's nonconformity score is atypical since the prediction is statistically unsupported by the holdout data, suggesting the test input is likely out-of-distribution and the prediction is unreliable. Conversely, a high $p$-value signifies that the test input aligns well with the holdout data, indicating the prediction is credible and supported by the data distribution.
The predicted beam label is the one with the highest $p$-value, as given by \begin{equation}
    \text{Prediction} = \operatorname*{argmax}_{j \in \left\{1,\dots,Q\right\}} p_j(\hat{\mathbf{x}}_S).
    \label{eq:prediction}
\end{equation}
Confidence is defined as one minus the second-highest $p$-value, i.e., the probability that any label other than the prediction is true, as given by
\begin{equation}
    \text{Confidence} = 1 - \max_{j \in \left\{1,\dots,Q\right\}, j \neq \text{prediction}} p_j(\hat{\mathbf{x}}_S).
    \label{eq:confidence}
\end{equation}
The prediction credibility is the $p$-value of the predicted beam label, which measures the degree to which the test input conforms to the training data, and is formulated as
\begin{equation}
    \text{Credibility} = \max_{j \in\left\{1,\dots,Q\right\}} p_j(\hat{\mathbf{x}}_S).
    \label{eq:credibility}
\end{equation}

\begin{algorithm}[t] \small
\caption{DkNN-based Robust Beam Assessment}
\label{DkNN_algo}
\KwIn{Training data $(X, Q) \in \mathcal{D}$, holdout data $ (X^c, Q^c)$, trained DNN $f^{'}\hspace{-0.15cm}\left(\mathbf{x}_S; {\boldsymbol{\theta}}\right)$ with $L$ layers, number of neighbors $\Tilde{k}$, test input $\hat{\mathbf{x}}_S$}
\For{$\eta = 1$ to $L$}{
    $\Pi \leftarrow \widetilde{k}$ points in ${X}$ closest to $\hat{\mathbf{x}_S}$ found using LSH tables\;
    $\Omega_\eta \leftarrow \left\{{q}_i : i \in \Pi\right\}$ 
    }
$\mathcal{C}$ =$ \left\{\varrho(\hat{\mathbf{x}}^c_u, {q}^c) : (\hat{\mathbf{x}}^c_u, q^c) \in ({X}^c, {Q}^c)\right\}$ \;
\For{$j = 1$ to $Q$}{
    $\varrho(\hat{\mathbf{x}}_S, j) \leftarrow \sum_{\eta = 1}^{L} \left|\left\{ i \in \Omega_\eta : i \neq j \right\}\right|$\;
    \BlankLine
    $p_j(\hat{\mathbf{x}}_S) \leftarrow \frac{|\{\varrho \in \mathcal{C} : \hspace{0.1cm}\varrho \geq \varrho(\hat{\mathbf{x}}_S, j)\}|}{|\mathcal{C}|}$
}
Calculate prediction, confidence, and credibility using equations \eqref{eq:prediction}, \eqref{eq:confidence}, and \eqref{eq:credibility}\;
\Return prediction, confidence, and credibility\;
\end{algorithm}

The proposed Adversarially-Enhanced DkNN-based Credible Beam Assessment Algorithm is summarized in Algorithm \ref{DkNN_algo}.{Note the the DkNN based beam prediction model is deployed at the BS and is capable of detecting outlier inputs  based on the assigned credibility scores.} {Unlike standard DNNs that treat the softmax output of the final layer as a confidence estimate, the DkNN approach provides predictions with both confidence and credibility scores. Confidence reflects the likelihood of the prediction being correct based on the training set, while credibility measures the relevance of the training set to the prediction, aiding in detecting outlier inputs.}

To evaluate the model’s resilience against
adversarial/outlier inputs, we generate the out-of-training
data (adversarial) examples by the Fast
Gradient Sign Method (FGSM) \cite{adv_FGSM} that aims
to manipulate the inputs to a beam classifier by perturbing them in the direction that maximizes the loss function in \ref{ref_loss} with respect to
the true beam labels. To generate the adversarial example, the FGSM  computes the perturbations as
$
\boldsymbol{\delta}_S = \epsilon \operatorname{sign}\left(\nabla_{\mathbf{x}_S} \mathcal{L}( \boldsymbol{\theta}, \mathcal{D})\right), $ 
where $\nabla$ is the gradient operator, and $\epsilon$ controls the perturbation magnitude under some suitable power
constraint with respect to the original RSSI values. The adversarial input computed as $
\mathbf{x}_{\mathrm{adv}, S} = \mathbf{x}_S + \boldsymbol{\delta}_S$ is used to assess the classifier's robustness to outlier inputs. 

To characterize the credibility estimates, we adopt the standard reliability diagrams \cite{RD}  to visualize the calibration of credibility scores. Reliability diagrams are histograms presenting accuracy as a function of credibility estimates of the model's prediction. The reliability diagrams bin the classifier's credibility score into $S$ intervals of equal size. 
A test data point $(\hat{\mathbf{x}}_S, q)$  is placed in bin $\mathcal{B}_s$ if the model’s credibility on $\hat{\mathbf{x}}_S$ is contained within the bin, i.e., $(\hat{\mathbf{x}}_S, q) \in \mathcal{B}_s$ . For each bin $\mathcal{B}_s$, the within-bin accuracy is defined as:
\begin{equation}
\operatorname{Acc}\left(\mathcal{B}_s\right)=\frac{1}{\left|\mathcal{B}_s\right|} \sum_{(\hat{\mathbf{x}}_S, q) \in \mathcal{B}_s} \mathbbm{1}_{\{\hspace{0.05cm} \operatorname{argmax}_{j \in \left\{1,\dots,Q\right\}} p_j(\hat{\mathbf{x}}_S) = {q}\}},
\end{equation}
which measures the fraction of test samples within the bin that are correctly classified.

It is noteworthy that rather than blindly trusting the model's predictions, the DkNN-based approach ensures consistency between the model's intermediate computations and its final output, improving explainability. The nearest neighbors in the DkNN approach offer example-based insights, with training points close to the test input providing relevant evidence, enabling a human observer to better understand and rationalize the beam prediction decisions.

\section{Simulation Results}\label{sec:simulation}
In this section, we provide details of the simulation setup, dataset acquisition, and the DL model architecture, followed by a discussion of the results.
\subsection{Simulation Setup}
\subsubsection{Scenario Setup}
 To simulate both the \textit{target scenario} (destination domain) and the \textit{DT scenario} (source domain), we utilize Wireless InSite software~\cite{remcom2024} to conduct precise {ray-tracing simulations}. Wireless InSite models the physical attributes of real-world objects, including their size, location, and material properties, to generate detailed {3D ray-tracing maps} of the communication environment. This software is employed for both scenarios to ensure consistent and realistic modeling of mmWave communications.

{In the \textit{target scenario}, representing a real-world deployment, we simulate mmWave BS-UE communications within an urban environment modeled after the downtown sector of Boston. The channel dataset is generated using the Boston5G scenario from DeepMIMO \cite{alkhateeb2019deepmimo}, which offers a high-fidelity simulated outdoor environment constructed with Wireless InSite's ray-tracing. This scenario includes a BS, a defined service area, and detailed representations of buildings and foliage, accurately capturing the propagation characteristics of an urban city landscape.}
The environment comprises both {line-of-sight (LOS)} and {non-line-of-sight (NLOS)} users. The BS employs a ULA with \( N_{\mathrm{BS}} = 32 \) antennas and is positioned at a height of \( 15 \, \mathrm{m} \), oriented towards the negative \( y \)-axis. UE devices are modeled as single-antenna systems at a height of \( 2 \, \mathrm{m} \). The service area spans \( 200 \, \mathrm{m} \times 230 \, \mathrm{m} \) and is discretized into a user grid with a spacing of \( 0.37 \, \mathrm{m} \). 
To enhance the stability and efficiency of training, the channel vectors \( \mathbf{h} \in \mathbb{C}^{N_{\mathrm{BS}} \times 1} \) are normalized by the largest absolute value in the channel matrix.

The \textit{DT scenario} (source domain) approximates the real-world environment through a \textit{digital replica}, which can be dynamically updated based on live monitoring of the environment and UE distribution. To generate the DT scenario, we adopt the approach described in ~\cite{alkhateeb2019deepmimo} and construct the channel matrices using the DeepMIMO dataset generator~\cite{deepmimo2024}. {Particularly, the DeepMIMO dataset generator is completely defined by 1) ray-tracing scenario and 2) dataset parameters. In the considered DT scenario setup, we adopt the ``Boston5G"  ray-tracing scenario ~\cite{alkhateeb2019deepmimo}. To account for the approximation errors in the EM 3D models and the difficulty of representing \textit{foliage} due to seasonal variations and unpredictable growth patterns, controlled imperfections are introduced into the DT scenario: {foliage} is completely omitted, and {building positions} are randomly shifted by \( 2 \, \mathrm{m} \) on the \( X \)-\( Y \) plane. The dataset parameter adhere to the parameters and system configurations used for the target scenario,} summarized in Table~\ref{tab:hyperparameters}.


\subsubsection{Dataset Acquisition}
The BS performs analog beamforming with $M_\mathrm{w}= 32$ sensing beams using a $N_{\mathrm{BS}}$-DFT codebook, whereas, for the narrow beam codebook $\mathbf{W}$, we use an O-DFT  with oversampling  factor of 4 to get a total of 128 narrow beams. The parameters of the neural network $\boldsymbol{\theta}$  are learned from the augmented dataset $\mathcal{D} = \mathcal{D}_\mathrm{R} \cup \mathcal{D}_\mathrm{T}$.
In all experiments, 70\% of the data is allocated for training, 10\% for validation, and the remaining 20\% for testing. The validation dataset serves as the holdout dataset and is also used for generating adversarial data, following the process described in Section \ref{Robust-BAE}, to evaluate the model's robustness.

All simulations are performed on a $10$-Core Intel(R) Xenon(R) Silver $4114$, $2.2 \mathrm{GHz}$ system equipped with an Nvidia Quadro P2000 graphics processing unit (GPU).

\begin{table}[!t]
\centering
\caption{\textsc{Hyperparameters for channel generation}}
\label{tab:hyperparameters}
\begin{adjustbox}{width=0.65\columnwidth,center}\footnotesize
\begin{tabular}{|c|c|}
\hline
\textbf{Name of scenario} & \textbf{Boston-5G} \\ \hline
Active BS          & 1               \\ \hline
BS transmit power          & 30 dBm               \\ \hline
Active users       & 1-622          \\ \hline
Number of antennas (x, y, z) & (1, 32, 1)  \\ \hline
Carrier frequency    & {28 GHz}        \\ \hline
System bandwidth          & 500 MHz        \\ \hline
Antenna spacing    & 0.5            \\ \hline
OFDM sub-carriers  & 1             \\ \hline
OFDM sampling factor & 1              \\ \hline
OFDM limit         & 1              \\ \hline
$D_\mathrm{T}, D_\mathrm{R},D_\mathrm{BG} $ size & 57144, 24485, 8162 \\ \hline
\end{tabular}
\end{adjustbox}
\end{table}

\begin{table}[!t] 
\centering
\caption{\textsc{Architecture and training hyperparameters for dnn and cnn models}}
\label{tab:Model-parameters}
\begin{adjustbox}{width=1\columnwidth,center}\small
\begin{tabular}{|c|c|c|c|} 
\hline
\textbf{Model}  & \textbf{Layer}  & \textbf{Details}   & \textbf{Activation} \\ \hline
\multirow{4}{*}{DNN} & Input Layer & Neurons = \# RSSI values & -  \\ \cline{2-4}
 & Hidden Layer 1   & 64 Neurons                      & ReLU    \\ \cline{2-4}
 & Hidden Layer 2   & 64 Neurons                      & ReLU    \\ \cline{2-4}
 & Hidden Layer 3   & 128 Neurons                     & ReLU    \\ \hline
\multirow{5}{*}{CNN} & Conv Layer 1        & Filters = 32, Kernel = $3\times3$, Stride = 1, Padding = 1 & ReLU                \\ \cline{2-4}
& Conv Layer 2        & Filters = 64, Kernel = $3\times3$, Stride = 1, Padding = 1 & ReLU                \\ \cline{2-4}
& Conv Layer 3        & Filters = 128, Kernel = $1\times1$, Stride = 1, Padding = 0 & ReLU                \\ \cline{2-4}
& Fully Connected     & 128 Neurons                     & -                   \\ \hline
\multicolumn{2}{|c|}{\textbf{Optimizer}}  & \multicolumn{2}{c|}{Adam, Learning rate = $10^{-3}$}          \\ \hline
\multicolumn{2}{|c|}{\textbf{Loss Function}} & \multicolumn{2}{c|}{Cross Entropy}                          \\ \hline
\multicolumn{2}{|c|}{\textbf{Epochs}}        & \multicolumn{2}{c|}{100}                                   \\ \hline
\end{tabular}
\end{adjustbox}
\end{table}

\subsection{Deep Learning Model Architecture}
To show that the proposed feature selection and the robustness assessment methodologies are model agnostic, we consider two different model architectures: 1) A convolutional neural network (CNN) based classifier with $L=$ 4 layers with three convolution layers stacked with a fully connected layer. Each convolution layer is followed by the rectified linear unit (ReLU) activation to provide non-linearity to the convolutional layers. 2) A fully connected DNN classifier with  $L=$ 3 fully connected layers and using ReLU activation, which takes the  RSSI values as input features, i.e., the input neurons equal to the number of RSSI values, followed by three hidden layers. The output layer has neurons corresponding to the possible O-DFT indices and outputs the softmax probability distribution. {For a quick nearest neighbor search on DkNN, we use  LSH from the FALCONN Python library, which implements cross-polytope LSH and offers sublinear query time performance \cite{andoni2015practical}. We employ a grid parameter search to configure the number of nearest neighbors to  $\widetilde{k}$ = $10$.}
The hyperparameters for the neural network architectures are summarized in Table \ref{tab:Model-parameters}.

\subsection{ Baselines and Metrics}
For comparison purposes, we consider the following benchmark methods:

\begin{enumerate}
\item \textit{SVD perfect channel} The upper bound beamforming based on the singular value
decomposition (SVD) of perfectly known channels under
$3$-bit quantized phase shifter \cite{OS_Asmaa}.  It is worth mentioning that the SVD upper bound
can only be reached when each user’s perfect channel is known at the BS.

\item \textit{Exhaustive search O-DFT} (with oversampling factor of $\times$4):   The BS exhaustively sweeps
the O-DFT codebook to scan $ N_{\mathrm{BS}} \times$4 narrow beams, and selects
the beam with the highest RSSI value for downlink transmission. 

{\item \textit{2-Tier Hierarchical Beam Search:} The BS employs a two-tier hierarchical codebook, where the tier-1 codebook comprises multiple wide beams, and the tier-2 codebook includes all the narrow beams from the adopted O-DFT codebook of size $Q$. Each narrow beam in the tier-2 codebook is  a child beam associated with a wide beam from the tier-1 codebook. The BS first sweeps the $M_\mathrm{w}=32$ wide beams from tier-1. It then sweeps the child beams in tier-2 that are associated with the best wide beam identified during the first step. The child beam with the highest received power is then selected. In this method, the wide beams are generated using the alternative minimization method with a closed form expression (AMCF) algorithm proposed in \cite{AMCF}. For $M_\mathrm{w}$ wide beams swept in tier-1, the number of narrow beams under each wide beam are $\lceil Q / M_\mathrm{w} \rceil$. The total number of beam measurements required is then given by  $M_\mathrm{w} +\lceil Q / M_\mathrm{w} \rceil$.}
{\item \textit{Binary Hierarchical Beam Search:} The BS performs a binary tree search on the narrow-beam DFT codebook of size $Q$. The search is organized into $\log_2 Q$ hierarchical layers. At each layer, the BS divides the current angular search space into two partitions and sweeps two corresponding wide beams generated using the AMCF technique. This process continues until a single narrow beam in $Q$ is identified.
}
{\item \textit{Fixed Beam Selection:} Based on the adaptation of the method proposed in \cite{jalali2024fast}, the BS  selects a fixed subset of RSSI values collected over a subset of narrow beams from the adopted $0$-DFT codebook  to train a CNN to associate the correct/best-oriented beam between the BS and
UE. The RSSIs over the fixed selected subset of narrow beams are used as the CNN’s
inputs and the index of the corresponding optimal narrow beam from the $0$-DFT codebook as the output. Note that this learning-based approach uses the same number of inputs as the proposed approach, but does not account for the importance of input features (RSSIs) when selecting the input subset for model training.}



\end{enumerate}
To evaluate the explainability and robustness of the proposed DkNN-based beam classifier, we compare its credibility estimates with the outputs of a standard softmax classifier, typically interpreted as
 model’s confidence estimates \cite{Deepknn}, with both classifiers using the same neural network architecture.

\subsection{Performance metrics}
We consider the following metrics for performance comparison:
\subsubsection{\textit{Top-k} accuracy} This measures the accuracy when the optimal beam index is among the top $k$ predicted beams. {Beam alignment is improved by sweeping the top $k$ beams with the highest predicted probabilities, enabling over-the-air refinement with $k$ additional overhead.}

\subsubsection{Beam Sweep Time $(T_\mathrm{sweep})$}
This is defined as the time taken to sweep a desired number of beams, and is linearly proportional to the number of beams swept. Typically, IA is performed periodically and constitutes $T_\mathrm{sweep}$ and the beam prediction time $T_\mathrm{predict}$, defined as the time it takes to process the
collected RSSI data and predict the beam it belongs to, such that the total IA time $T_\mathrm{IA}= T_\mathrm{sweep} +T_\mathrm{predict}$. While $T_\mathrm{predict}$ depends on the processing capability at the BS, $T_\mathrm{sweep}= N_bt_s$ depends on the number of candidate beams scanned $N_b$, where $N_b= \widetilde{M}_{\mathrm{w}}$, and $t_s$  is the time required to scan a single candidate beam \cite{FastIA}. Generally, $T_\mathrm{sweep} \gg T_\mathrm{predict}$, and therefore we consider the beam sweep time for comparing different schemes.

\begin{figure}[!t]
    \centering
    \includegraphics[width=0.9\linewidth]{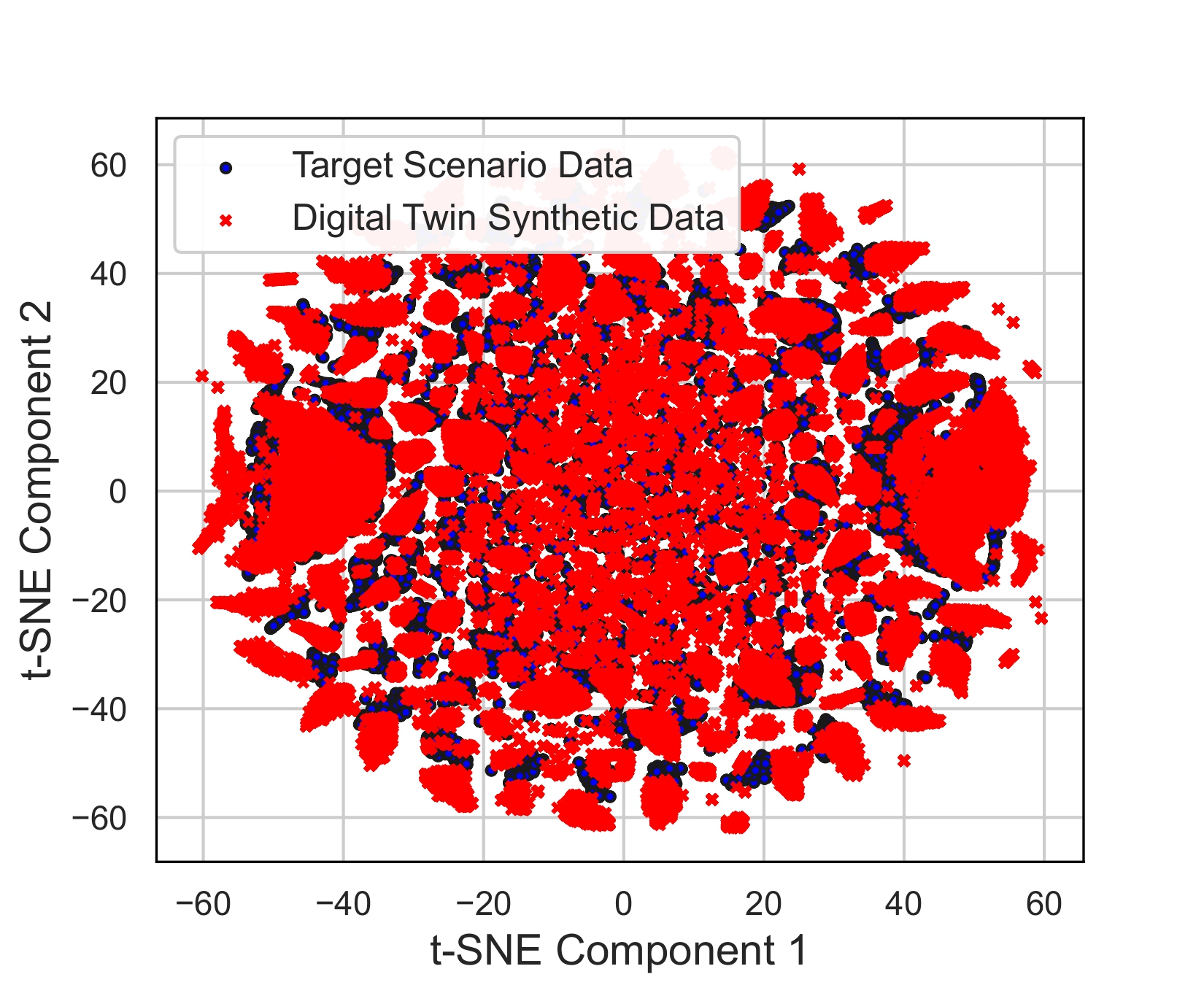}
    \caption{T-SNE visualization of the generated data from the target scenario and digital twin scenario.}
    \label{fig:t-SNE}
\end{figure}

\subsubsection{Effective Spectral Efficiency} This quantifies the achievable rate of a communication system while accounting for the overhead associated with beam alignment. It is defined as the proportion of channel resources dedicated to data transmission after subtracting the time spent on beam alignment during the channel coherence period. Mathematically, it is expressed as:

\begin{equation} \mathrm{SE} = \frac{T_\mathrm{frame} - T_\mathrm{IA}}{T_\mathrm{frame}} \log_2\left(1 + \mathrm{SNR}\right), \end{equation}
where $T_\mathrm{frame}$ represents the duration of one frame during which the channel response remains fixed, i.e., the total time for a communication session \cite{deep-tl}.

\subsection{Performance Evaluation}

{\subsubsection{T-SNE Visualization} Fig. \ref{fig:t-SNE} shows a t-distributed stochastic neighbor embedding
(t-SNE) visualization of data points from the target and DT scenario. The goal of t-SNE is to reduce the dimensionality
and visualize high-dimensional data in a two-dimensional
(2D) plane, preserving the structure and relationships between
points. It achieves this by minimizing the Kullback-Leibler (KL) divergence between the joint probability distributions in the original and reduced spaces. The figure reveals a relatively uniform scatter with a central clustering, indicating that the high-dimensional data has been effectively projected into a lower-dimensional space by t-SNE, while preserving local neighborhood structures and relationships. Since the two datasets share the same layout
with slight modifications, the data distributions exhibit similarity despite the introduction of minor imperfections in the DT-aided synthetic data.}

\subsubsection{Analysis of XAI-based Input Feature Selection}
 \begin{figure}[t]
    \centering
    \includegraphics[width=1\linewidth]{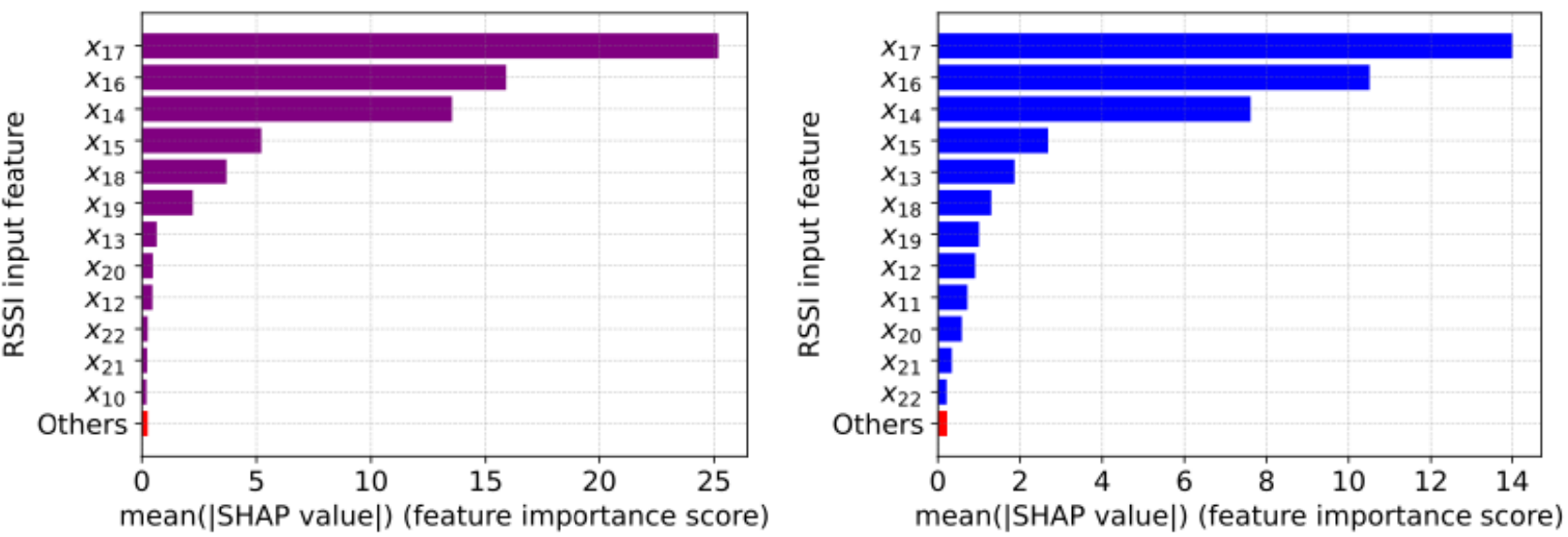}
    \setlength{\belowcaptionskip}{15pt}
    \caption{Feature importance rankings for the beam classifier generated using CNN (left) and fully connected DNN (right).}
    \label{fig:shap_bar_plot}
\end{figure}

We  arrange the obtained feature importance scores into rankings as shown
in Fig. \ref{fig:shap_bar_plot}. By analyzing these rankings, we can identify the
most and the least important features for the DL-based BAE decisions. The horizontal axis shows the average impact on
output action, and the vertical axis shows the most influential input features in decreasing order of their magnitude. From Fig. \ref{fig:shap_bar_plot}, it can be observed that the top most essential features selected using Algorithm \ref{algo_fixed} are mostly in agreement irrespective of the CNN/DNN based model architecture.{ For visualization purposes, we display the top $12$ selected features, while the combined effect of the remaining features is represented as ``Others."} Further, to select the subset of important features, we set the threshold \(\delta\) at 71\%, 82\%, 92\%, 96\%, and 99\%, corresponding to the selection of $\widetilde{M}_{\mathrm{w}}=2, 4, 8, 12$, and $24$ important features, respectively.

Next, we analyze the performance of the DNN models trained on subsets of importance-based input features, i.e., the RSSIs over selected sensing beams. While similar analyses apply to CNN-based beam prediction, initial experiments revealed that the pre-trained DNN outperforms CNN in $\textit{ Top-}k$ accuracy on the site-specific dataset, leading to the selection of the DNN-based BAE for further evaluation. Moreover, we denote the model trained on synthetic/augmented data as the DNN-TL-synthetic and the model purely trained on real-world data as DNN-real, respectively.

\subsubsection{Analysis of Model Refinement using Transfer Learning}Fig. \ref{fig:aug} demonstrates the \textit{top}-$2$ prediction accuracy versus the importance-based features selected using Algorithm \ref{algo_fixed}.  
The DNN-TL-synthetic model is initially pre-trained on $57,144$ synthetic data points generated from the DT scenario, where the building positions include a modeling error of $2$ meters. In addition, a separate real-world dataset comprising $24,485$ data samples is used to train a dedicated DNN-real model. A subset of the real-world dataset is further utilized to fine-tune the DNN-TL-synthetic model with varying augmentation sizes.  For a given number of importance-based selected features, a dedicated DNN-TL-synthetic model is trained, which requires an average training time of approximately $3$ minutes per configuration.

\begin{figure}[t]
    \centering
    \includegraphics[width=1\linewidth]{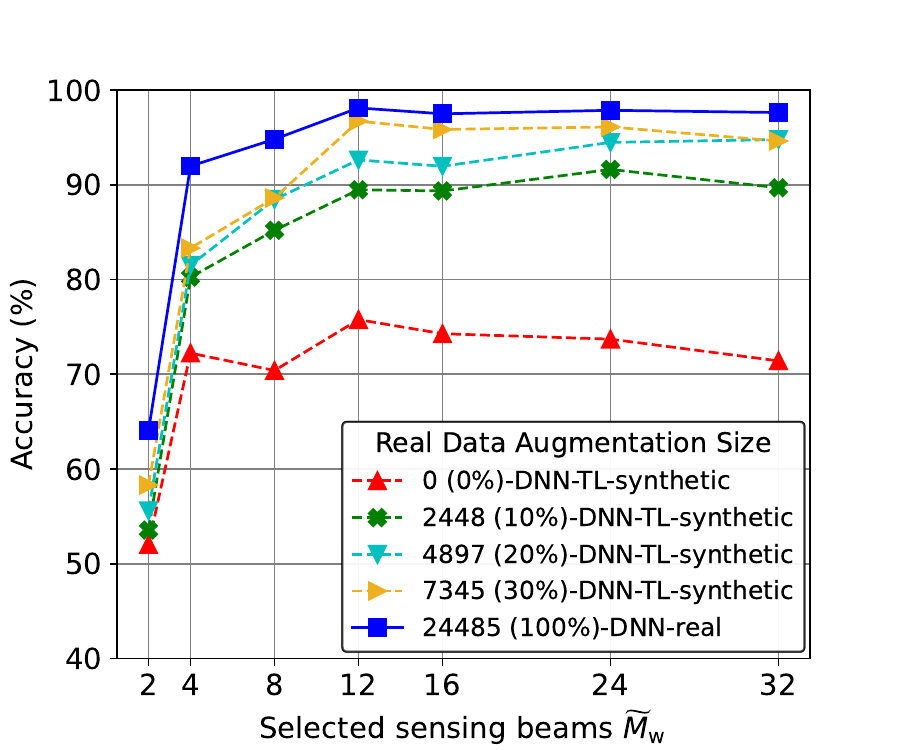}
    \caption{Top-2 accuracy vs. number of selected sensing beams/features. The DNN-TL-synthetic model is trained using synthetic data samples from the DT environment and fine-tuned using real-world data augmentation.}
    \label{fig:aug}
\end{figure}

{Fig. \ref{fig:aug} illustrates that  evaluating  the DNN-TL-synthetic model trained purely on synthetic data performs poorly when tested on the unseen target data.} By training with limited real data augmentation, e.g, $10\%$ of the destination domain dataset $\mathcal{D}_{\mathrm{R}}$, yields a notable improvement in beam alignment accuracy compared to the scenario where no samples captured at the destination domain are used for model fine-tuning. The figure highlights that with 30\% real data augmentation, the DNN-TL-synthetic model, leveraging RSSI values over $\widetilde{M}_{\mathrm{w}}=12$ importance-based features, attains approximately 98\% of the accuracy performance of the DNN-real model trained on the complete real-world dataset. Notably, the DNN-TL-synthetic model achieves this performance while requiring 70\% fewer real-world data samples, showcasing model refinement through transfer learning effectively reduces the data collection overhead in real-world systems. For the remaining simulation results, we consider the DNN-TL-synthetic model fine-tuned with only 30\% of the real-world data samples, as it efficiently calibrates mismatches between real-world model and digital replica with minimal real-world data requirement.



\begin{figure}[t]
    \centering
    \includegraphics[width=1\linewidth]{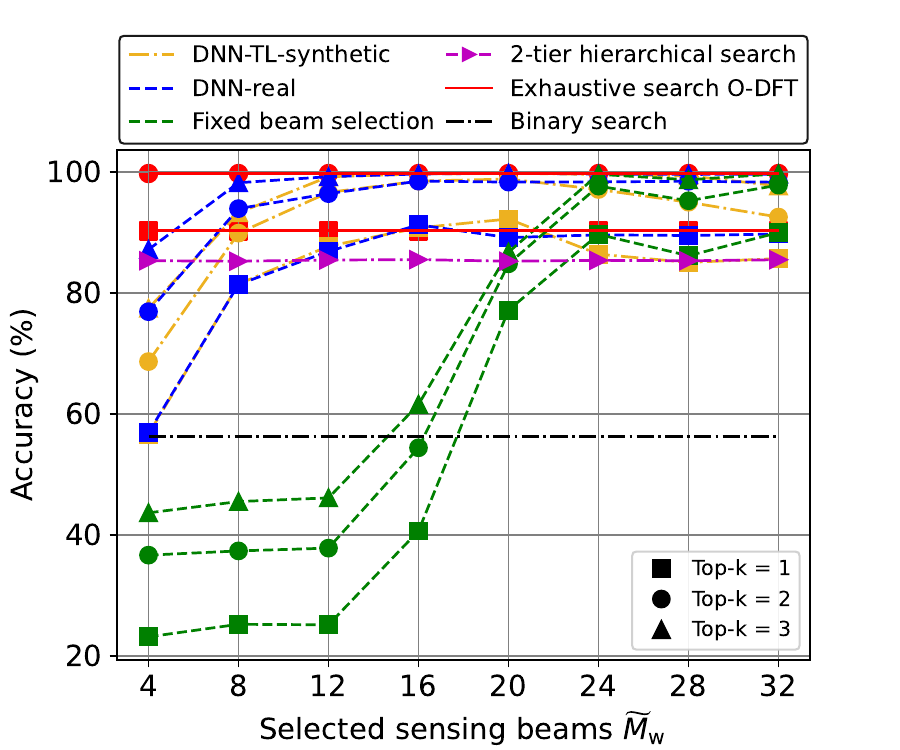}
    \caption{Top-k accuracy performance of different schemes under measurement noise.}
    \label{fig:accuracy_vs_beams}
\end{figure}

\subsubsection{Analysis of Beam Alignment Accuracy under Measurement Noise} The accuracy of beam prediction depends on the measured power of beamforming signals, where noise in these signals can degrade beam alignment accuracy.  The  DNN-TL-synthetic model is trained with no measurement noise present in its
training data, but it is then exposed to noisy signal measurements during the
testing/validation stage. In our analysis, the noise
power ranges considered while generating the noisy signal are
between -$94\mathrm{dBm}$ and -$114\mathrm{dBm}$, and randomly sampled to generate the unseen target data. For a fair comparison, we ensure that all schemes consider the same noise level.

Fig. \ref{fig:accuracy_vs_beams} shows the \textit{top-k =}$\{1, 2, 3\}$  accuracy against the number of importance-based  features $\widetilde{M}_{\mathrm{w}}$. The exhaustive search achieves the highest accuracy due to its low susceptibility to noise in beam measurements. However, this accuracy comes at the expense of excessive training overhead and occasional search errors caused by the induced noise, resulting in beam alignment mismatches. {The binary search performs the worst among the traditional beam sweeping-based baselines since the binary  search  introduces more layers and is thus more prone to error propagation; a suboptimal decision in any upper layer may lead to an incorrect final beam selection due to noise or imperfect beam patterns. The importance of feature selection becomes evident from the performance of the fixed beam search approach, which treats all features/beams as equally important. It requires sweeping at least 24 sensing beams to achieve an accuracy of  87\%.} The proposed method can achieve a beam alignment accuracy of 87\% using measurements from $\widetilde{M}_{\mathrm{w}}=12$ importance-based sensing beams and can attain 95\% accuracy by searching an
additional $k = 2$ narrow beams. With  $\widetilde{M}_{\mathrm{w}} = 12$ important features, the fine-tuned model achieves \textit{top-3} accuracy of $\approx 97\%$ outperforming the exhaustive search method. This improvement is mainly attributed to eliminating less important and non-informative RSSI inputs, which would complicate the learning process with noisy measurements. 

\subsubsection{Analysis of Average SNR} The beam alignment accuracy considers the probability of
finding the optimal narrow beam. With a large O-DFT
codebook, adjacent narrow beams may have similar beamforming gains. As a result, operators may be more interested in the average SNR achieved after beam alignment.

\begin{figure}[t]
    \centering
    \includegraphics[width=1\linewidth]{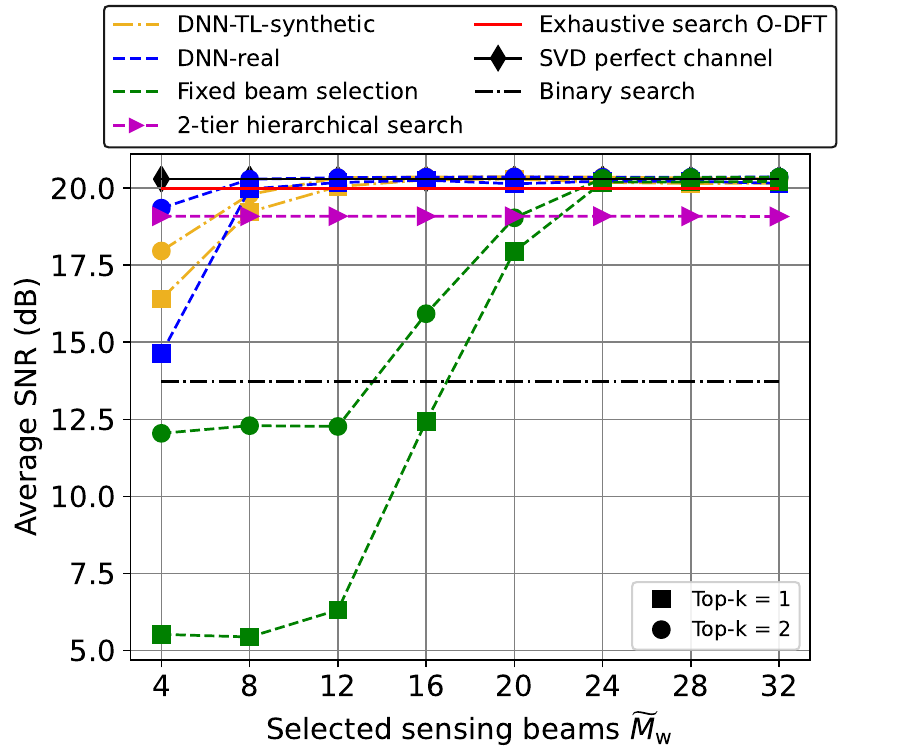}
    \caption{Average SNR versus the number of selected sensing beams/features for different schemes.}
    \label{fig:SNR_vs_features}
\end{figure}

Fig. \ref{fig:SNR_vs_features} illustrates the average SNR achieved by the proposed method and the benchmark schemes with the increasing number of selected features $\widetilde{M}_{\mathrm{w}}$. The exhaustive search achieves close-to-optimal average SNR of the SVD-based solution as it searches over all possible beams in the O-DFT codebook. {The DNN-TL-synthetic model fine-tuned with 30\% real-world data, remarkably matches the average SNR of the DNN-real model trained on the entire real dataset, using just $12$ out of $32$ importance-based sensing beams  and outperforms all traditional benchmarks including the exhaustive search. This is due to the fact that the DL methods can also learn the statistical property of noise, which is lacking in conventional
methods. Furthermore, the proposed SHAP-based beam selection method reduces the beam sweeping overhead by 50\% compared to the fixed beam search approach, which does not account for the relevance of input features in the prediction process. The degraded performance of fixed beam search approach may be attributed to the  fact that the best beam may be overlooked since it might not be included in
the measured set during training.   The superior performance of the proposed DNN-based BAE stems from the SHAP-based feature selection approach, which identifies the ``quality of information" in these selected subsets of features to be more valuable and sufficient for the model in identifying the narrow beams best oriented to the UEs.} Furthermore, the proposed DNN-TL-synthetic model delivers performance comparable to the exhaustive search solution with $\approx$ 89\% reduction in beam training overhead compared to the exhaustive search, demonstrating its effectiveness in challenging NLOS environments.

\subsubsection{Analysis of Effective Spectral Efficiency} Fig. \ref{fig:SE_vs_features} shows the achievable effective spectral efficiency using the proposed and benchmark schemes across different number of selected features $\widetilde{M}_{\mathrm{w}}$. Contrary to the SVD-based solution and the exhaustive search approach, where the number of input features is fixed at $32$ and $128$, respectively, the proposed solution adapts the candidate sensing beams $\widetilde{M}_{\mathrm{w}}$ based on the selected features determined by their feature importance. With  $\widetilde{M}_{\mathrm{w}}=8$  beams, the proposed DNN-TL-synthetic model can outperform the SVD-based solution in terms of effective spectral efficiency metric and delivers the same performance as the DNN-real model by searching an
additional $k = 2$ narrow beams. The exhaustive search method requires scanning $128$ narrow beams and has the worst spectral effectiveness despite having superior average SNR performance. { The hierarchical search and binary tree search based  approaches require  $M_\mathrm{w} +\lceil Q / M_\mathrm{w} \rceil$ and $2+ 2\log_2 (Q/2)$ beam searches for $M_\mathrm{w}$ wide beams and $Q$ narrow beams, respectively, and therefore exhibit degraded effective spectral efficiency. } An interesting observation is that the effective spectral efficiency has a unique maximizer $\widetilde{M}^{*}_{\mathrm{w}}$: it
increases initially with $\widetilde{M}_{\mathrm{w}} \leq$$\widetilde{M}^{*}_{\mathrm{w}}$ as the beam alignment probability improves with  $\widetilde{M}_{\mathrm{w}}$. However, as $\widetilde{M}_{\mathrm{w}} $ increases beyond $\widetilde{M}^{*}_{\mathrm{w}}$, this
gain is offset by the increased overhead and reduced time
for the data communication.

\begin{figure}[t]
    \centering
    \includegraphics[width=1\linewidth]{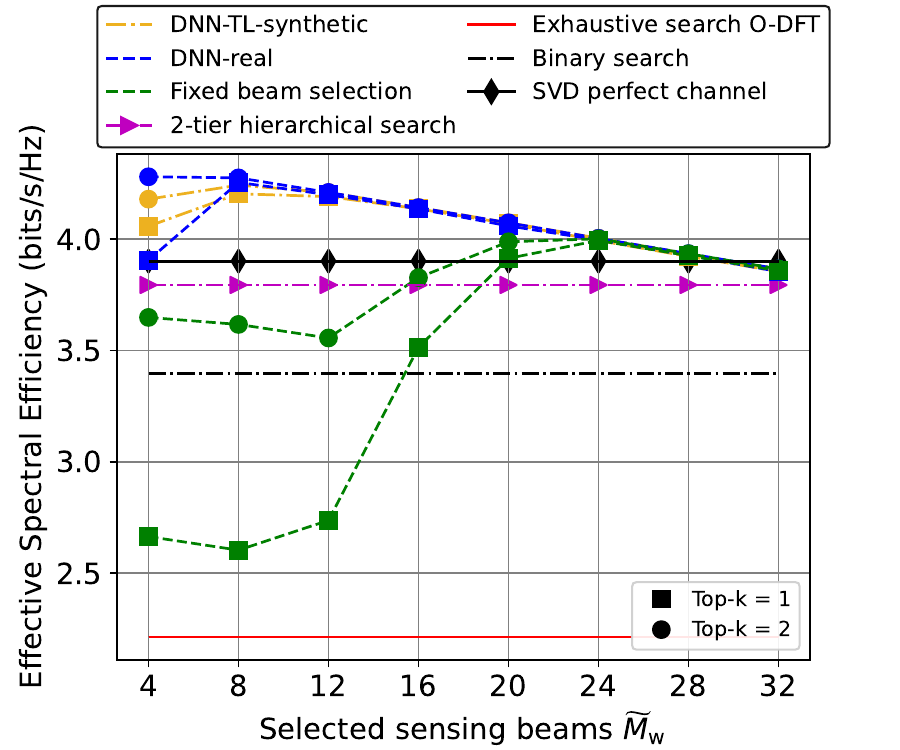}
    \caption{Effective spectral efficiency vs. number of selected sensing beams/features for different approaches.}
    \label{fig:SE_vs_features}
\end{figure}

\subsubsection{ Computation, Model and Time complexity}
{The computation complexity of the proposed DL-based BAE is determined by the beam sweeping complexity and feedback complexity, and is
linearly proportional to the number of beams swept. With the proposed DL-based BAE , all $N_\mathrm{U}$ UEs can measure and report the RSSIs over the $
\widetilde{M}_{\mathrm{w}}$ sensing beams  simultaneously when the BS
sweeps the selected sensing beams. If the BS chooses to sweep the
top-$k$ predicted beams, the selected beams could vary across different UEs. Hence, the {beam sweeping complexity} is $
\widetilde{M}_{\mathrm{w}} + N_\mathrm{U} \times k \cdot\mathbbm{1}_{\{k > 1\}}
$, whereas the {feedback complexity} will be $N_\mathrm{U} \times \widetilde{M}_{\mathrm{w}} + N_\mathrm{U} \cdot \mathbbm{1}_{\{k > 1\}} $, since if the BS chooses to sweep additional beams, each UE only needs to feedback the index of the best beam.}
The model complexity of the proposed DL-based BAE  depends on the structure of implemented DNNs (number of hidden layers and the number of corresponding neurons). {The model complexity is primarily dominated by the update process of the network weight parameters. Let $H_l$ denote the number of neurons in hidden layer $L$.  The total number of scalar parameters is $
\mathcal{O}\left( |\mathbf{x}_S| \cdot H_1 + \sum_{l=1}^3 H_l \cdot H_{l+1} \right) $, where $|\mathbf{x}_S| =\widetilde{M}_{\mathrm{w}}$ is the input size, i.e, the number of input neurons.  The update process of the loss function is linear to the mini-batch size $B$ since the optimizer scans all mini-batch samples. As a result, an update of all parameters incurs a computational complexity of $
\mathcal{O}\left(B \cdot \left( |\mathbf{x}_S| \cdot H_1 + \sum_{l=1}^3 H_l \cdot H_{l+1} \right)\right) $}.

Table \ref{tab:model_parameters} shows the complexity
comparison in terms of the trainable parameters, accuracy, and IA time for different numbers of selected sensing beams $\widetilde{M}_{\mathrm{w}} $. From the table, it is evident that the beam sweep time dominates the overall IA time. For instance, 5G NR  specifies a duration  $5\mathrm{ms}$ for sweeping across a total of $64$ sensing beams. Based on this, the allowed scanning time per beam is \(   0.078 \, \mathrm{ms} \)  \cite{tutorialBA}, \cite{FastIA}. Compared to the exhaustive search method requiring $\approx 10 \mathrm{ms}$ for sweeping across $128$ narrow beams and predicting the optimal beam, the proposed DL-based BAE finishes beam sweeping in $\approx 0.98 \mathrm{ms}$, using just $12$ sensing beam while maintaining close to exhaustive search accuracy. Overall, using just $12$  importance-based features out of full set of $32$ features/sensing beams, the proposed DL-based BAE can reduce the trainable parameters by $\approx4.29\%$, and the beam sweeping time by $\approx  90.5\%$ compared to the exhaustive search method, respectively. 

{ The prediction times for proposed DL-aided BAE and the learning and non-learning based schemes are tabulated in Table \ref{tab:timing}}. {Note that the prediction times depend on the system's computational resources. The computational capabilities of BSs in practical networks typically surpass those of the machines used in our simulations. As a result, the average time required for training and evaluating beam alignment models, the SHAP based feature selection and model robustness assessment can be further reduced in real-world deployments.}

\begin{table}[!t]
\centering
\caption{\textsc{Model and time complexity for proposed beam alignment method }}
\label{tab:model_parameters}
\begin{adjustbox}{width=1\columnwidth,center}
\begin{tabular}{|c|c|c|c|}
\hline
 \multicolumn{1}{|c|}{\makecell{\textbf{No. of Sensing} \\ \textbf{ Beams/Features $\widetilde{M}_{\mathrm{w}} $}}} & \multicolumn{1}{c|}{\makecell{\textbf{Trainable} \\ \textbf{Parameters}}} & \multicolumn{1}{c|}{\makecell{\textbf{Top-1/Top-2/Top-3} \\ \textbf{Accuracy (\%)}}} & \makecell{\textbf{IA Time (ms)} \\ \textbf{$T_\mathrm{IA} = T_\mathrm{sweep} + T_\mathrm{predict}$}} \\ \hline
2  & 29,184 & 44 / 55 / 63 & 0.15625 + 0.0427 = 0.19895 \\ \hline
4  & 29,312 & 56 / 68 / 77 & 0.3125 + 0.0428 = 0.3553 \\ \hline
8  & 29,568 & 81 / 88 / 93 & 0.625 + 0.0425 = 0.6675 \\ \hline
12 & 29,824 & 87 / 96 / 97 & 0.9375 + 0.0435 = 0.9850 \\ \hline
16 & 30,080 & 90 / 98 / 99 & 1.25 + 0.0414 = 1.2914 \\ \hline
24 & 30,592 & 87 / 97 / 99 & 1.875 + 0.0478 = 1.9228 \\ \hline
32 & 31,104 & 88 / 93 / 98 & 2.5 + 0.0416 = 2.5416 \\ \hline

\end{tabular}
\end{adjustbox}
\end{table}

\begin{table}[ht]
\centering
\caption{\textsc{Prediction time} (in ms) \textsc{for  different algorithms}}
\label{tab:timing}
\resizebox{1\columnwidth}{!}{
\begin{tabular}{|c|c|c|c|}
\hline
\makecell{\textbf{No. of Sensing} \\ \textbf{Beams/Features} $\widetilde{M}_{\mathrm{w}}$} & 
\makecell{\textbf{DNN-TL-synthetic} \\ \textbf{(proposed)}} & 
\makecell{\textbf{DNN-Real} \\ \textbf{}} & 
\makecell{\textbf{Fixed Beam} \\ \textbf{Selection}} \\
\hline
4  & 0.0428 & 0.0429 & 0.0865 \\
\hline
8  & 0.0427 & 0.0456 & 0.0500 \\
\hline
12 & 0.0475 & 0.0484 & 0.0490 \\
\hline
16 & 0.0414 & 0.0786 & 0.0465 \\
\hline
20 & 0.0466 & 0.0479 & 0.0460 \\
\hline
24 & 0.0478 & 0.0432 & 0.0461 \\
\hline
28 & 0.0745 & 0.0427 & 0.0489 \\
\hline
32 & 0.0416 & 0.0486 & 0.0507 \\
\hline
\multicolumn{4}{|c|}{\textbf{Non-learning Schemes with Fixed Number of Beam Measurements}} \\
\hline
\makecell{Exhaustive Search  (128 beams)}           & \multicolumn{3}{c|}{0.0591} \\
\hline
\makecell{2-tier Hierarchical  Search (36 beams)$^\dagger$} & \multicolumn{3}{c|}{1.7110} \\
\hline
\makecell{Binary Search  (14 beams)$^\star$} & \multicolumn{3}{c|}{0.9519} \\
\hline
\end{tabular}
}
\vspace{-1mm}
\begin{flushleft}
\footnotesize
$^\dagger$ 32 wide beam sweeps + 4 narrow sweeps per wide beam. \\
$^\star$ Computed as $2 + 2\log_2(128/2)$ narrow beam sweeps.
\end{flushleft}
\end{table}

{The proposed DL model training and the post-
hoc SHAP-based feature selection is site-specific and conducted offline assisted by the DT. After training, a
single simplified model using the selected features is deployed for beam prediction at that site.  If the channel statistics change significantly  due to model drift or seasonal variations, the retraining of the deployed model and update of feature selection would be necessary. In such scenarios, the DNN may either be trained from scratch or initialized with existing weights and fine-tuned using data from the new environment. Retraining is triggered when beam alignment performance (e.g., Top-k accuracy or L1-RSRP difference) falls below a threshold. Since large-scale environmental properties tend to vary gradually, retraining and feature selection are anticipated to be needed only occasionally.}

\subsubsection{Robustness Evaluation} We evaluate
the robustness of the proposed DL-based BAE to adversarial inputs by incorporating the  DkNN-based approach described in Section \ref{Robust-BAE}, and comparing its prediction credibility to that of the softmax classifier. The softmax classifier estimates confidence using output probability distributions but lacks robustness against out-of-distribution/ adversarial inputs \cite{Deepknn}. In contrast, incorporating the DkNN approach into the proposed beam classifier provides interpretability through nearest neighbors, offering human-understandable explanations for intermediate computations at
each layer, making it a valuable debugging tool. We show that
the softmax classifier is poorly calibrated and overestimates
confidence when predicting out-of-distribution inputs.

Fig. \ref{fig:RD1} and Fig. \ref{fig:RD2} present the reliability diagrams for the DkNN and the naive softmax classifier, respectively. The distribution of credibility/confidence values across the data is given by the number of data points in each credibility bin, reflected by the red line overlaid on the bars. The softmax classifier lacks calibration as it consistently exhibits high confidence on both test and adversarial data, making it ineffective in identifying outliers.  Fig. \ref{fig:RD1} demonstrates that the proposed DkNN classifier exhibits superior calibration by assigning low credibility to adversarial samples, effectively filtering outliers. It achieves $8.5 \times$ and $3 \times$ robustness improvements at credibility thresholds of $0.2$ and $0.4$, respectively. Fig. \ref{fig:RD2} illustrates the softmax classifier reliability diagram, which shows overconfidence, misclassifying more than 70\% of adversarial examples with high confidence ($>0.9$). Note that because of the NLOS environment in the Boston-5G dataset,  the test dataset contains many test inputs that are
difficult to classify, reflected by the lower mean accuracy of the
underlying DNN. Still, the DkNN recovers some accuracy on adversarial examples by leveraging representations from DNN's internal layers and, therefore, is better calibrated than its softmax equivalent.

\begin{figure*}[t]
    \begin{subfigure}[t]{0.45\linewidth}
        \centering
        \includegraphics[width=\linewidth]{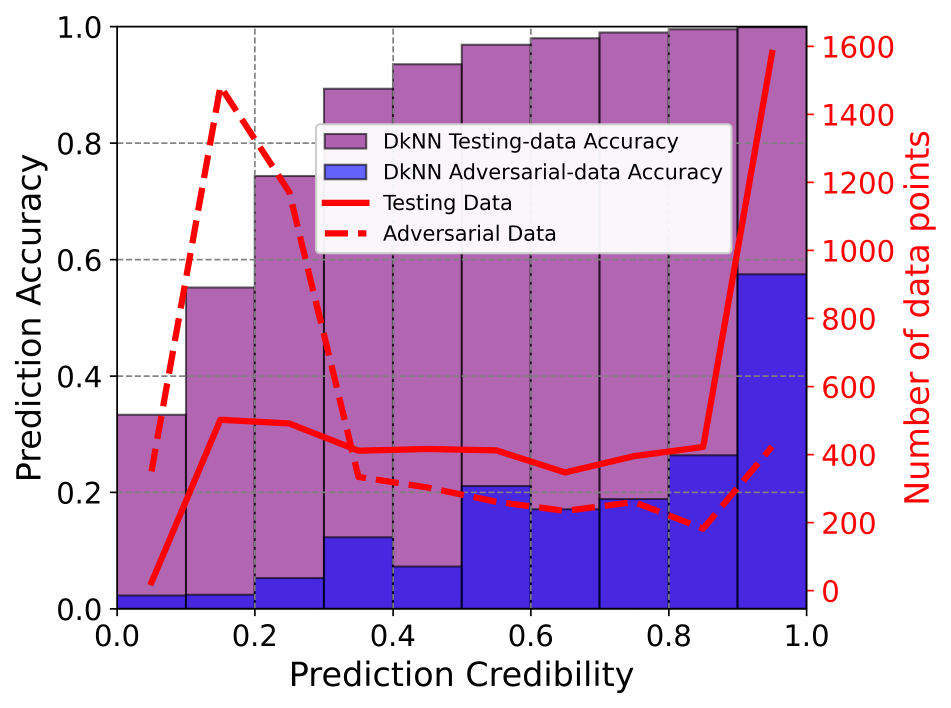}
        \caption{}
        \label{fig:RD1}
    \end{subfigure}
    \hfill
    \begin{subfigure}[t]{0.45\linewidth}
        \centering
          \includegraphics[width=\linewidth]{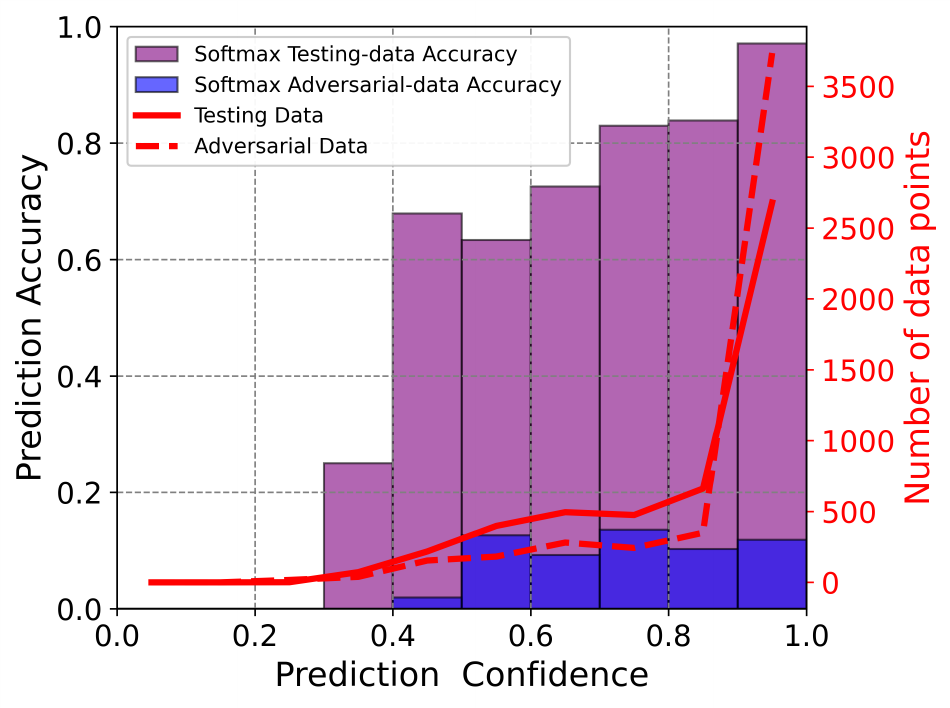}
        \caption{}
        \label{fig:RD2}
    \end{subfigure}
    \setlength{\belowcaptionskip}{-10pt}
    \caption{ (a) Reliability diagram for DkNN Classifier, (b) Reliability diagram for Softmax Classifier.}
    \label{fig:RD}
\end{figure*}

\section{Conclusion}\label{sec:conclusion}
In this paper, we have developed a robust and explainable DL-based BAE for mmWave MIMO systems that uses RSSI measurements from wide sensing beams to efficiently predict optimal narrow beams. By generating synthetic data with a site-specific DT and refining the pre-trained model through transfer learning, we reduce real-world data requirements by up to 70\%. We leverage the model-agnosticism and explainability power of SHAP to design an efficient sensing beam selection strategy to further reduce the beam training overhead by prioritizing critical features and cutting beam training overhead by at least $62$\% compared to the model trained on the full feature set while preserving near-optimal spectral efficiency and beam alignment accuracy. Furthermore, we integrate the DkNN algorithm into the designed DL-based BAE to enhance robustness against out-of-distribution inputs by up to $8.5$$\times$ compared to the traditional softmax-based classifier. The current approach for transfer learning stems from domain adaptation with site-specific characteristics. Future works could involve considering task adaptation, where trained DNN weights can be effectively
reused in other propagation environments or antenna
configurations by fine-tuning them with small datasets in the
destination environment to improve generalization. 

\bibliographystyle{IEEEtran}\bibliography{IEEE_abr,XAI_bibliography}

\end{document}